\documentclass[sigconf]{acmart}

\usepackage{amsmath,amsfonts}

\usepackage{graphicx}
\usepackage{threeparttable}
\usepackage{textcomp}
\usepackage{xcolor}
\usepackage{tabularx}
\colorlet{shadecolor}{gray!40}
\usepackage{tcolorbox}  

\usepackage{caption}
\usepackage{subcaption}
\usepackage[title,titletoc]{appendix}

\usepackage{algorithm} 
\usepackage{algpseudocode}
\usepackage{float}

\usepackage[T1]{fontenc}
\usepackage{booktabs}
\usepackage{multirow}
\usepackage{rotating}
\usepackage{amsmath}
\usepackage{float}

\usepackage[table]{xcolor}
\usepackage{threeparttable} 

\usepackage{hyperref}

\usepackage{bm}
\def\BibTeX{{\rm B\kern-.05em{\sc i\kern-.025em b}\kern-.08em
    T\kern-.1667em\lower.7ex\hbox{E}\kern-.125emX}}

\usepackage[table]{xcolor}
\usepackage{multirow}
\usepackage{xfp}
\usepackage{ifthen}

\newcommand{\AQE}{AdaQE-CG}

\AtBeginDocument{%
  \providecommand\BibTeX{{%
    Bib\TeX}}}

\copyrightyear{2026}
\acmYear{2026}
\setcopyright{cc}
\setcctype{by}
\acmConference[WWW '26]{Proceedings of the ACM Web Conference 2026}{April 13--17, 2026}{Dubai, United Arab Emirates}
\acmBooktitle{Proceedings of the ACM Web Conference 2026 (WWW '26), April 13--17, 2026, Dubai, United Arab Emirates}
\acmPrice{}
\acmDOI{10.1145/3774904.3792685}
\acmISBN{979-8-4007-2307-0/2026/04}

\ccsdesc[500]{Information systems~Information retrieval}
\ccsdesc[500]{Computing methodologies~Natural language processing}
\ccsdesc[500]{Applied computing~Document management and text processing}

\keywords{Web-scale Document, Generative AI, Model Card, Data Card, Large Language Model, Query Expansion}
\begin{document}

\title{AdaQE-CG: Adaptive Query Expansion for Web-Scale Generative AI Model and Data Card Generation}

\author{Haoxuan Zhang}
\email{haoxuanzhang@my.unt.edu}
\affiliation{%
  \institution{University of North Texas}
  \city{Denton}
  \state{TX}
  \country{USA}
}

\author{Ruochi Li}
\email{rli14@ncsu.edu}
\affiliation{%
  \institution{North Carolina State University}
  \city{Raleigh}
  \state{NC}
  \country{USA}
}

\author{Zhenni Liang}
\email{zhenniliang@my.unt.edu}
\affiliation{%
  \institution{University of North Texas}
  \city{Denton}
  \state{TX}
  \country{USA}
}

\author{Mehri Sattari}
\email{mehrisattari@my.unt.edu}
\affiliation{%
  \institution{University of North Texas}
  \city{Denton}
  \state{TX}
  \country{USA}
}

\author{Phat Vo}
\email{phatvo@my.unt.edu}
\affiliation{%
  \institution{University of North Texas}
  \city{Denton}
  \state{TX}
  \country{USA}
}

\author{Collin Qu}
\email{collinqu@gmail.com}
\affiliation{%
  \institution{Bellevue High School}
  \city{Bellevue}
  \state{WA}
  \country{USA}
}

\author{Ting Xiao}
\email{ting.xiao@unt.edu}
\affiliation{%
  \institution{University of North Texas}
  \city{Denton}
  \state{TX}
  \country{USA}
}

\author{Junhua Ding}
\email{junhua.ding@unt.edu}
\affiliation{%
  \institution{University of North Texas}
  \city{Denton}
  \state{TX}
  \country{USA}
}

\author{Yang Zhang}
\authornote{Corresponding authors.}
\email{yang.zhang@unt.edu}
\affiliation{%
  \institution{University of North Texas}
  \city{Denton}
  \state{TX}
  \country{USA}
}

\author{Haihua Chen}
\authornotemark[1]
\email{haihua.chen@unt.edu}
\affiliation{%
  \institution{University of North Texas}
  \city{Denton}
  \state{TX}
  \country{USA}
}

\renewcommand{\shortauthors}{Haoxuan Zhang et al.}

\begin{abstract}
Transparent and standardized documentation is essential for building trustworthy generative AI (GAI) systems. However, current automated model and data card generation methods still face three key challenges: \textbf{(i) Static templates.} Most systems rely on fixed query templates that cannot adapt to diverse paper structures or evolving documentation requirements. \textbf{(ii) Information scarcity.} Web-scale repositories such as Hugging Face often provide incomplete or inconsistent metadata, resulting in missing or noisy information. \textbf{(iii) Lack of benchmarks.} The absence of standardized datasets and evaluation protocols prevents fair and reproducible assessment of documentation quality. To address these challenges, we propose \textbf{AdaQE-CG}, an \textit{Adaptive Query Expansion for Card Generation} framework that integrates dynamic information extraction with cross-card knowledge transfer. The \textbf{Intra-Paper Extraction via Context-Aware Query Expansion (IPE-QE)} module iteratively refines extraction queries to capture richer and more complete information from scientific papers and repositories. The \textbf{Inter-Card Completion using the MetaGAI Pool (ICC-MP)} module enriches missing fields by transferring semantically relevant content from similar cards within a curated dataset. In addition, we construct \textbf{MetaGAI-Bench}, the first large-scale, expert-annotated benchmark for evaluating GAI documentation. Comprehensive experiments across five quality dimensions demonstrate that AdaQE-CG significantly outperforms existing approaches, surpasses human-authored data cards, and approaches human-level quality for model cards. Code, prompts, and data are publicly available at: \url{https://github.com/haoxuan-unt2024/AdaQE-CG}.
\end{abstract}

\maketitle

\section{Introduction}

As artificial intelligence (AI) systems proliferate across the web at an unprecedented scale, the need for transparent and standardized documentation has become increasingly critical. Model and data cards, which are fundamentally web-based data artifacts, have emerged as key transparency mechanisms that provide structured, machine-readable documentation for models and datasets underpinning AI systems \cite{mitchell2019model, pushkarna2022data}. In the era of generative AI (GAI), where large language models (LLMs), multimodal architectures, and web-scale GAI platforms exhibit complex behaviors and opaque data dependencies, these web-native documentation frameworks are indispensable for conveying information about training corpora, performance characteristics, and potential biases \cite{guo2024bias, semmelrock2025reproducibility, mihalcea2025ai}. 

Beyond promoting transparency, model and data cards constitute a critical layer of web infrastructure for enabling responsible, reproducible, and scalable GAI development—particularly in high-stakes domains such as healthcare, finance, and law, where provenance, fairness, and ethical accountability are essential \cite{gilbert2025could, raza2025responsible}. As web-scale GAI ecosystems continue to expand, characterized by emergent model behaviors, intricate data supply chains, and dynamic deployment contexts, model and data cards provide the minimal yet indispensable accountability substrate that supports development, evaluation, governance, and interoperability across the global AI landscape.

Model and data card generation also forms the foundation for a broad spectrum of critical applications that advance transparency, accountability, and governance in AI systems. These web-based documentation artifacts enable longitudinal tracking of model and dataset evolution, supporting systematic analyses of changes in architecture, data composition, and performance over time \cite{laufer2025anatomy, liang2024systematic, yang2024navigating}. They also power public repositories and registries that facilitate cross-model and cross-dataset comparisons of capabilities, biases, licenses, and usage constraints, thereby promoting interoperability and reproducibility within the broader AI ecosystem \cite{heming2023benchmarking, bhardwaj2024state, sokol2024benchmarkcards}. Moreover, model and data cards are instrumental in benchmarking and governance, providing structured metadata for automated compliance auditing against ethical principles, data consent requirements, and security standards \cite{longpre2024data, puhlfurss2025model}. Beyond compliance, they serve as key enablers of risk and fairness auditing, helping researchers and practitioners identify, quantify, and mitigate potential harms in AI deployment contexts \cite{longpre2024large}. Collectively, these applications highlight the indispensable role of model and data cards as the connective infrastructure that bridges AI research, practice, and policy across web-scale generative systems. 

\begin{figure}[!t]
    \centering 
    \includegraphics[width=0.47\textwidth]{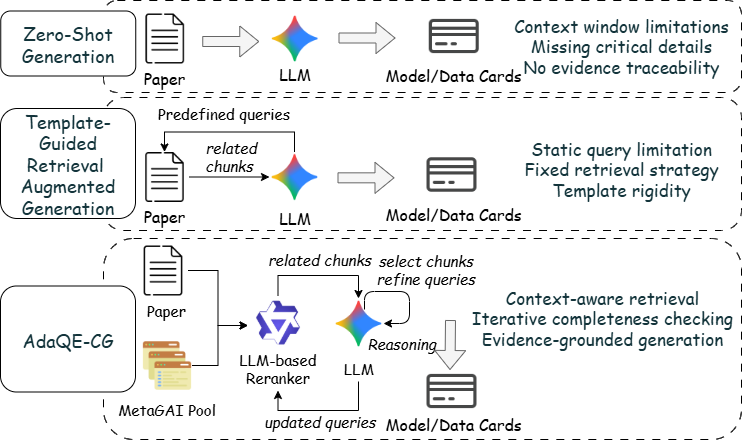}
    \caption{A comparison of card generation paradigms: (a) Zero-Shot Generation (top), (b) Template-Guided RAG (middle), and (c) \AQE \space (bottom).} 
    \label{method-diff} 
\end{figure}

However, model and data card generation is a non-trivial task. Manual card creation faces significant scalability challenges, as current human-generated documentation suffers from inconsistencies, incompleteness, and a heavy reliance on developers' subjective interpretations of what should be reported \cite{yang2024navigating, liang2024systematic,liu2024automatic}. Automatic generation of model and data cards promises consistency and scalability, but faces several challenges, as shown in Figure~\ref{method-diff}. Zero-shot generation synthesizes complete documentation in a single inference pass but encounters three critical deficiencies: context window constraints that truncate lengthy papers, missing critical details due to a lack of systematic extraction, and the absence of evidence traceability that prevents source verification. Template-guided RAG methods \cite{liu2024automatic} employ predefined question templates yet face distinct limitations: static query constraints preventing adaptation to diverse paper structures, fixed retrieval strategies failing to accommodate domain-specific patterns, and template rigidity creating schema mismatches with evolving documentation requirements. Beyond paradigm-specific constraints, both approaches confront shared challenges: (1) the absence of ground truth makes it difficult to objectively assess completeness and correctness, (2) algorithms that rely solely on information from academic papers or web-based data sources such as Hugging Face or GitHub rarely achieve optimal performance due to information scarcity in these sources, (3) the risk of hallucinations when LLMs summarize lengthy documentation beyond their context windows, and (4) inconsistent templates and schema variants across data sources.

To address these limitations, we propose \textbf{\AQE}, an \textbf{Ada}ptive \textbf{Q}uery \textbf{E}xpansion for \textbf{C}ard \textbf{G}eneration framework that tackles both static query constraints and information scarcity. The first module, Intra-Paper Extraction via Context-Aware Query Expansion (IPE-QE) dynamically adapts extraction strategies by iteratively refining queries based on identified information gaps, overcoming static template limitations. The second module, Inter-Card Completion using the MetaGAI Pool (ICC-MP) enriches incomplete fields by transferring knowledge from architecturally and semantically similar cards in the curated MetaGAI Pool. This hybrid approach balances automation efficiency with documentation quality while maintaining provenance transparency. 

In a nutshell, our contributions are summarized as follows:

\begin{itemize}
    \item \textbf{MetaGAI: Web-Scale Dataset and High-Quality Benchmark.} 
    We construct the MetaGAI-Dataset of 6,481 data cards and 123,013 model cards across four GAI modalities, and establish a high-quality MetaGAI-Bench of 1,200 expert-annotated cards with strong inter-annotator agreement. 
    
    \item \textbf{Comprehensive Empirical Statistical Analysis.} 
    We introduce the Weighted Card Completeness Index (WCCI) to quantify documentation quality. Our analysis reveals that data cards significantly outperform model cards, with critical gaps in responsible AI fields. We identify strong correlations between completeness and popularity through systematic correlation analysis.

    \item \textbf{\AQE: Adaptive Query Expansion Framework.} 
    We propose a novel hybrid-module architecture combining IPE-QE for dynamic extraction and ICC-MP for cross-card knowledge transfer, and validate its effectiveness through comprehensive ablation studies.

    \item \textbf{Rigorous Evaluation Framework and Empirical Validation.} 
    We employ a multi-perspective assessment combining LLM-as-a-Judge and human evaluation across five quality dimensions (\textit{Faithfulness, Relevance, Accuracy, Consistency, Usefulness}). Our \AQE \space framework outperforms the human baseline for data cards and achieves human-level performance for model cards in LLM evaluation. In human evaluation, our framework consistently ranks second among all methods.
\end{itemize}

\section{Related Work}
\label{related}

\textbf{Web-scale semantic integration} transforms heterogeneous documentation into standardized, machine-interpretable knowledge. For example, an analysis of 2 million models on Hugging Face (a large web repository) revealed declining documentation quality as model cards become templated and auto-generated \cite{laufer2025anatomy}. Horwitz et al. \cite{horwitz2025charting} further mapped over 400,000 model relationships to infer undocumented attributes, underscoring the need for scalable semantic synthesis. To improve coverage and consistency, Liu et al. \cite{liu2024automatic} proposed CardGen, building a 4.8k model and 1.4k data card corpus. Beyond repositories, large-scale efforts such as AutoSchemaKG \cite{bai2025autoschemakg}, TEXT2DB \cite{jiao2024text2db}, structured scientific extraction using LLMs \cite{dagdelen2024structured}, and retrieval-augmented generation for multi-document and multimodal reasoning \cite{liu2024automatic, suri2025visdom}, further demonstrate the trend toward progressive web-scale synthesis of structured knowledge.

\textbf{Model and data card generation is also critical for responsible AI} as accountability and traceability become major concerns in web-scale GAI applications \cite{papagiannidis2025responsible, novelli2024accountability, kroll2021outlining}. A lack of visibility into how models are trained, evaluated, and deployed often obscures their limitations and societal risks, underscoring the need for transparent documentation \cite{panch2019artificial, daneshjou2021lack, jacovi2021formalizing}. For models, Model Cards \cite{mitchell2019model} were proposed to disclose evaluations across demographics and usage conditions, and subsequent research expanded their scope with explainability principles \cite{phillips2021four}, consumer-style labels \cite{seifert2019towards}, complementary card families \cite{adkins2022prescriptive, shen2021value}, and toolkits for tracking and reporting model information \cite{arya2019one}. More recently, this line of work has evolved toward regulatory compliance and system-level governance \cite{brajovic2023model}, as well as end-to-end transparency and trust \cite{sidhpurwala2025blueprints}. For data, Datasheets \cite{gebru2021datasheets}, Data Statements \cite{bender2018data}, and Data Nutrition Labels \cite{holland2020dataset} set baseline practices for recording provenance, consent, and fitness-for-use. These foundations were succeeded by templated Data Cards and accompanying guidance for documenting crowd-sourced data \cite{pushkarna2022data}. Building on this line of work, recent studies have explored automated generation of dataset documentation \cite{liu2024automatic} and the design of machine-readable open datasheets \cite{roman2023open, rondina2023completeness} to improve completeness, consistency, and discoverability in large-scale repositories. Despite these advances, prior methods remain largely static, relying on predefined templates or constrained multi-source extraction.

\section{Preliminary}

\label{task}

Model cards and data cards provide structured, machine-interpretable documentation for AI models and datasets. Following Mitchell et al. \cite{mitchell2019model} and Pushkarna et al. \cite{pushkarna2022data}, we define a semantic card as $C = \{f_1, f_2, \ldots, f_m\}$ where $m$ is the total number of fields, and each field $f_i = (k_i, v_i)$ comprises (1) a field name $k_i$ from the card taxonomy and (2) a field value $v_i$ containing natural language descriptions or structured data.

Given document chunks $\mathcal{D} = \{c_1, c_2, \ldots, c_n\}$ from a scientific paper and its web repository page (where $n$ is the total number of chunks), together with a \textbf{MetaGAI Pool} $\mathcal{M} = \{C_1, C_2, \ldots, C_N\}$, which is a curated collection of $N$ high-quality cards filtered from the MetaGAI-Dataset by completeness ($\tau_{wcci}$) and popularity ($\tau_{pop}$) criteria, the goal is to generate a complete card $C$ by extracting and synthesizing information across these sources.

\textbf{Task Definition:} The automated card generation task is formulated as two sequential modules:

\textbf{Module 1: Intra-Paper Extraction via Context-Aware Query Expansion (IPE-QE).} For each target field $f_{id}$, iteratively refine queries $q_r$ (round $r$) to extract information from chunks $\mathcal{D}$, terminating when information gain $\Delta_r$ falls below the threshold $\epsilon$ or completeness is reached. This produces an initial card $C'$:
\begin{equation}
\label{eq1}
C'[f_{id}] = A_{r^*}, \quad r^* = \min\{r : \Delta_r \leq \epsilon \lor \text{IsComplete}(A_r)\}
\end{equation}
where $A_r$ represents the generated answer for round $r$, computed as $A_r = \text{LLM}(q_r, \text{Retrieve}(\mathcal{D}, q_r))$, and $q_r$ is generated based on previous answers and query history $\mathcal{Q}$.

\textbf{Module 2: Inter-Card Completion using the MetaGAI Pool (ICC-MP).} Given the initial card $C'$ from Module 1, identify incomplete fields $\mathcal{F}_{inc} = \{f_i \in C' : v_i = \emptyset\}$ and retrieve similar cards from the MetaGAI Pool $\mathcal{M}$ via TF-IDF (Term Frequency–Inverse Document Frequency) tag matching (threshold $\alpha$) and semantic reranking (top-$k$). The enriched card $C$ is produced by:
\begin{equation}
\label{eq2}
C[k_i] = \text{Synthesize}(\mathcal{V}_i, C'), \quad \mathcal{V}_i = \{v_j : C_j[k_i] \in \mathcal{M}_{sim}\}
\end{equation}
where $\mathcal{M}_{sim} \subseteq \mathcal{M}$ is the set of top-$k$ most relevant cards, $\mathcal{V}_i$ represents the collection of field values for field name $k_i$ from similar cards, and $C$ represents the final enriched card. This module is applied only when $\mathcal{F}_{inc} \neq \emptyset$ and similar cards exist in $\mathcal{M}$; otherwise, $C = C'$. Detailed algorithms for both modules are presented in Section~\ref{method}.

\section{MetaGAI Dataset and Benchmark}
\label{data}
\subsection{MetaGAI-Dataset Construction}

We construct the MetaGAI-Dataset from Hugging Face\footnote{Hugging Face: \url{https://huggingface.co}}, a prominent web-based platform hosting over 2 million open-source AI models and datasets. While the platform provides highly credible human-authored descriptions, these resources exhibit substantial heterogeneity and data sparsity, which pose significant challenges for semantic web applications. The descriptions lack standardized schemas and machine-interpretable semantics, and numerous fields remain unfilled, impeding automated knowledge integration.

To address these challenges, we implement a three-stage semantic data pipeline. First, we leverage the Hugging Face Hub API\footnote{Hugging Face Hub API: \url{https://huggingface.co/docs/huggingface_hub/}} to systematically retrieve models and datasets across four GAI modalities: Multimodal, Computer Vision, Natural Language Processing, and Audio. Second, we establish publication provenance by filtering resources with associated scientific papers through metadata tags containing DOI or arXiv identifiers, enabling paper-to-card semantic alignment. Third, we employ Gemini-2.5 Flash-Lite \cite{comanici2025gemini} to transform unstructured web descriptions into machine-interpretable JSON-formatted semantic cards following our standardized taxonomy (Table~\ref{tab:definition} in Appendix \ref{sec:definition}). Our semantic annotation process generates structured cards where each field is augmented with confidence scores $\{0.25, 0.5, 0.75, 1.0\}$ to support quality assessment and downstream completeness analysis. The resulting MetaGAI-Dataset comprises 6,481 data cards and 123,013 model cards with machine-interpretable semantics, collected on August 20, 2025. 

\subsection{MetaGAI-Bench Construction}
\label{sec:benchmark}
To evaluate the performance of the card generation algorithm, we randomly sampled 600 model cards and 600 data cards from the MetaGAI-Dataset for manual annotation to establish a high-quality benchmark. Our annotation team comprised six members: two Ph.D. students in Information Science and four master's students in Data Science, all with expertise in GAI. The team was divided into three groups. Annotators reviewed original Hugging Face content, associated papers, and generated structured cards to complete their annotations.

The annotation process followed two stages. In the first stage, each group independently annotated 50 model cards and 50 data cards. We then assessed intra-group consistency to ensure annotation quality. Traditional inter-annotator agreement metrics like Cohen's kappa \cite{cohen1960coefficient} are unsuitable for generative tasks, as identical meanings can be expressed through different wording without definitive correct answers. Therefore, we employed BERTScore \cite{zhangbertscore} to calculate semantic similarity for each card field between annotators, treating annotations as consistent when semantic similarity exceeded 0.8. This approach yielded average kappa-equivalent scores of 0.939 for model cards and 0.806 for data cards, both indicating high agreement. After resolving inconsistencies within each group, the second stage began, in which each group completed the remaining annotations.

\subsection{Statistics and Analysis}

To quantify documentation quality across heterogeneous card structures, we propose the Weighted Card Completeness Index (WCCI), a novel metric built upon three fundamental principles: interpretability, confidence-weighted evaluation, and uniform field weighting. The detailed formula and field-level analysis by task category are presented in Appendices \ref{wcci} and \ref{appendix:wcci_analysis}, respectively. To examine factors associated with card completeness, we computed Spearman correlations between WCCI scores and metadata tags, as shown in Table~\ref{tab:correlations} in Appendix~\ref{wcci}.

Popularity metrics demonstrate significant positive associations with documentation completeness for both types of cards, with model cards exhibiting particularly strong correlations compared to data cards. These positive correlations align with prior findings that more popular artifacts on Hugging Face consistently exhibit higher documentation quality~\cite{yang2024navigating, liang2024systematic}.

Task-specific patterns reveal an important paradox. Specialized tasks such as text-to-speech ($\rho = 0.211$) and multimodal synthesis ($\rho = 0.132$) show positive correlations with WCCI scores. However, text generation models show a significant negative correlation ($\rho = -0.153$), despite being the most common task category ($n=97{,}689$). This finding indicates that high-volume domains have lower documentation quality, creating gaps where standardized practices are most needed~\cite{yang2024navigating, liang2024systematic}. For data cards, NLP-centric tasks (zero-shot classification: $\rho = 0.203$, text classification: $\rho = 0.201$) show moderate positive correlations, reflecting more established documentation practices in these research areas.

For model cards, infrastructure and licensing choices show strong associations with documentation completeness. Apache-2.0 license ($\rho = 0.374$) and PyTorch framework ($\rho = 0.386$) have the strongest positive correlations, suggesting that open-source ecosystems and popular frameworks promote better documentation through community standards. For data cards, crowdsourced annotation shows a strong negative correlation with documentation completeness ($\rho = -0.384$), likely reflecting coordination challenges and a lack of centralized oversight.

\section{Methodology}
\label{method}

\begin{figure*}[!h]
    \centering 
    \includegraphics[width=\textwidth]{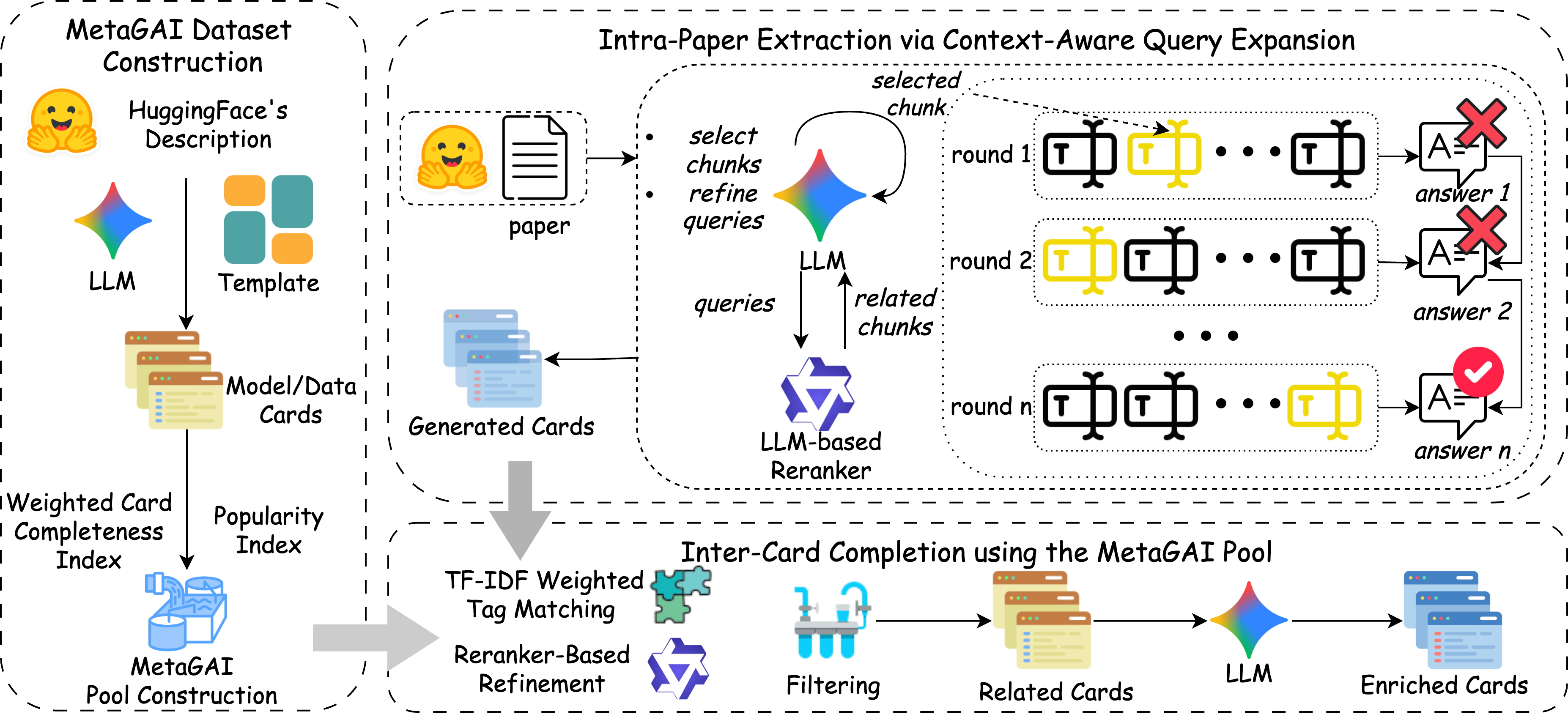}
    \caption{\AQE: Hybrid-module pipeline for automated card generation. Module 1 (IPE-QE): iteratively extracts field values from document chunks via context-aware query expansion to produce the initial card. Module 2 (ICC-MP): conditionally enriches incomplete fields using the MetaGAI Pool when similar cards are available, producing the final card.} 
    \label{framework} 
\end{figure*}

We propose \AQE, a hybrid-module framework that implements the pipeline defined in Section \ref{task} to generate complete semantic cards from academic papers and web repository metadata. As illustrated in Figure \ref{framework}, Module 1 (IPE-QE) iteratively extracts field values $v_i$ from document chunks $\mathcal{D}$ using dynamic query refinement and LLM-based reranking to produce the initial card $C'$. Module 2 (ICC-MP) is conditionally applied when $\mathcal{F}_{inc} \neq \emptyset$ and similar cards exist in the MetaGAI Pool $\mathcal{M}$, enriching incomplete fields through two-phase architectural and semantic matching to yield the final card $C$. This hybrid pipeline ensures deep extraction from source documents while enabling broad completion through cross-card knowledge transfer when applicable.

\subsection{Module 1: Intra-Paper Extraction via Context-Aware Query Expansion}

Module 1 implements the extraction process formalized in Equation \ref{eq1}. For each target field, we iteratively refine queries to extract field values from document chunks, terminating when information gain falls below the threshold or completeness is achieved.

\subsubsection{Document Preprocessing and Chunking}

To construct the document set $\mathcal{D}$, we parse scientific papers from PDF to Markdown using Nougat \cite{blecher2023nougat}. Each section of the paper forms an independent chunk $c_i$, while Hugging Face metadata constitutes a separate chunk. This section-level granularity balances information completeness with retrieval precision, ensuring each chunk $c_i \in \mathcal{D}$ contains semantically coherent content for effective extraction.

\subsubsection{LLM-Based Reranking for Retrieval}

We employ an LLM-based reranker \cite{qwen3embedding} to identify relevant chunks for each query. This approach enables contextual understanding of semantic relevance and effective handling of complex, multifaceted queries. The reranker produces ranked document chunks and filters them to retain only relevant information for answer generation.

\subsubsection{Dynamic Multi-Round Query Expansion}

As formalized in Algorithm \ref{alg:ipe-qe}, for each target field with its initial query, we generate an initial answer using the top-ranked relevant chunks from the reranker, then employ two adaptive mechanisms: \textit{context-aware query refinement} and \textit{adaptive stopping criterion}. After each round, the LLM evaluates answer completeness and generates refined queries by analyzing information gaps in the previous answer while leveraging accumulated query history to avoid redundant extraction. The \textit{ComputeGain} function assigns scores (0-3) based on completeness, quality, new information, and utility criteria, while \textit{IsComplete} evaluates whether the query generator indicates completion (returns ``COMPLETE''), signaling that all required information has been extracted. The expansion terminates when: (1) the answer is complete, (2) information gain is insufficient ($\leq \epsilon$) for two consecutive rounds, or (3) maximum rounds $R_{max}$ is reached, producing initial card $C'$ with extracted field values for all target fields.

\begin{algorithm}[htbp]
\small
\caption{Intra-Paper Extraction via Context-Aware Query Expansion (IPE-QE)}
\label{alg:ipe-qe}
\begin{algorithmic}[1]
\Require 
    \Statex $\mathcal{D} = \{c_1, c_2, \ldots, c_n\}$: Document chunks
    \Statex $\{f_1, \ldots, f_m\}$: Target fields where $f_i = (k_i, v_i)$
    \Statex $q_0$: Initial queries
    \Statex $R_{max}$, $\epsilon$: Maximum rounds (default: 10) and gain threshold (default: 1)
\Ensure 
    \Statex $C' = \{f_1, \ldots, f_m\}$: Initial card

\State $C' \gets \emptyset$

\For{each field $f_{id}$}
    \State $\mathcal{Q} \gets \{q_0^{(id)}\}$, $r \gets 0$, $\text{stall\_count} \gets 0$

    \While{$r < R_{max}$}
        \State $A_r \gets \text{LLM}(q_r, \text{RerankerRetrieve}(\mathcal{D}, q_r))$
        
        \If{$\text{IsComplete}(A_r, q_0^{(id)}, f_{id})$}
            \State \textbf{break} \Comment{Success: all information extracted}
        \EndIf
        
        \If{$r > 0$ \textbf{and} $\text{ComputeGain}(A_{r-1}, A_r, q_0^{(id)}, f_{id}) \leq \epsilon$}
            \State $\text{stall\_count} \gets \text{stall\_count} + 1$
        \Else
            \State $\text{stall\_count} \gets 0$
        \EndIf
        
        \If{$\text{stall\_count} \geq 2$}
            \State \textbf{break} \Comment{Stalled: insufficient progress}
        \EndIf
        
        \State $q_{r+1} \gets \text{RefineQuery}(q_0^{(id)}, A_r, \mathcal{Q}, f_{id})$
        \State $\mathcal{Q} \gets \mathcal{Q} \cup \{q_{r+1}\}$, $r \gets r + 1$
    \EndWhile
    
    \State $C'[f_{id}] \gets A_r$
\EndFor

\State \Return $C'$
\end{algorithmic}
\end{algorithm}

\subsection{Module 2: Inter-Card Completion using the MetaGAI Pool}

Module 2 implements the conditional enrichment process formalized in Equation \ref{eq2}. Given the initial card from Module 1, we identify incomplete fields and enrich them using similar cards from the MetaGAI Pool when available. If no incomplete fields exist or no similar cards are found, the initial card is returned as the final output.

The MetaGAI Pool $\mathcal{M}$ is pre-constructed from the MetaGAI-Dataset using dual quality criteria to ensure reliable knowledge sources: (1) \textit{Completeness Filtering} retains cards meeting WCCI thresholds (top 30\% for data cards and top 10\% for model cards) to ensure comprehensive documentation quality; (2) \textit{Popularity Filtering} prioritizes cards exceeding minimum download thresholds, reflecting community validation and practical utility \cite{liang2024systematic,yang2024navigating}. Our correlation analysis confirms this relationship in the MetaGAI-Dataset (Table \ref{tab:correlations}). This dual-criterion approach yields a curated pool combining thorough documentation with proven practical value.

For each incomplete field, we employ a two-phase retrieval pipeline followed by category-aware synthesis, as formalized in Algorithm \ref{alg:icc-mp}: 

\textit{Phase 1 - TF-IDF Tag Matching:} We compute TF-IDF weights for metadata tags such as task type and model architecture to identify architecturally similar cards. This structural filtering efficiently narrows the search space by retaining only cards exceeding a weighted overlap threshold $\alpha$, focusing on models with similar architectures or datasets with comparable characteristics. \textit{Phase 2 - Semantic Reranking:} An LLM-based reranker evaluates fine-grained semantic similarity between the target card and filtered candidates from Phase 1. This phase selects the top-$k$ most semantically relevant cards, capturing nuanced similarities beyond surface-level tag matching. The two-phase design balances computational efficiency with retrieval accuracy.

\textit{Category-Aware Synthesis:} We collect field values from retrieved similar cards and synthesize appropriate content based on field categorization. Fields are classified into three categories: (1) \textit{Shared Properties} such as intended use and limitations that can be reasonably inferred from architecturally similar cards, (2) \textit{Unique Properties} such as model developers and contact details that are artifact-specific and must never be transferred, and (3) \textit{General Information} such as ethical considerations that are broadly applicable across similar contexts. The LLM evaluates applicability for each category and synthesizes content that integrates relevant information while preserving artifact-specific accuracy.

\begin{algorithm}[htbp]
\small
\caption{Inter-Card Completion using the MetaGAI Pool (ICC-MP)}
\label{alg:icc-mp}
\begin{algorithmic}[1]
\Require 
    \Statex $C' = \{f_1, \ldots, f_m\}$: Initial card where $f_i = (k_i, v_i)$
    \Statex $\mathcal{M}$: MetaGAI Pool (pre-filtered by $\tau_{wcci}$ and $\tau_{pop}$)
    \Statex $k$, $\alpha$: Top-$k$ parameter (default: 10) and tag overlap threshold (default: 0.5)
\Ensure 
    \Statex $C$: Final card (enriched if similar cards found, otherwise $C = C'$)

\State $\mathcal{F}_{inc} \gets \{f_i \in C' : v_i = \emptyset\}$

\If{$\mathcal{F}_{inc} = \emptyset$}
    \State \Return $C'$
\EndIf

\State $C \gets C'$

\For{each field $f_i = (k_i, v_i) \in \mathcal{F}_{inc}$}
    \State $\mathcal{M}_{tag} \gets \{C_j \in \mathcal{M} : \text{TagOverlap}(C', C_j) > \alpha\}$
    \If{$\mathcal{M}_{tag} = \emptyset$}
        \State \textbf{continue}
    \EndIf
    
    \State $\mathcal{M}_{sim} \gets \text{RerankTopK}(C', \mathcal{M}_{tag}, k)$
    \State $\mathcal{V}_i \gets \{C_j[k_i] : C_j \in \mathcal{M}_{sim} \land C_j[k_i] \neq \emptyset\}$
    
    \If{$\mathcal{V}_i \neq \emptyset$}
        \State $C[k_i] \gets \text{Synthesize}(\mathcal{V}_i, C')$
    \EndIf
\EndFor

\State \Return $C$
\end{algorithmic}
\end{algorithm}

\section{Experiments and Results}
\label{exp}

In this section, we investigate three research questions to guide our experimental evaluation:

\begin{itemize}
    \item[\textbf{RQ1:}] How can web-scale information be effectively integrated to create comprehensive model and data cards for GAI?
    
    \item[\textbf{RQ2:}] How can we overcome static template limitations and leverage web-scale data to achieve high-quality card generation?
    
    \item[\textbf{RQ3:}] What evaluation frameworks can effectively assess the quality of generated cards across multiple quality dimensions?
\end{itemize}

\subsection{Experimental Setup}
\subsubsection{Baselines}
To evaluate the effectiveness of our method, we compare against two automated baselines and human-annotated cards as a reference standard, representing different paradigms in card generation.

\textbf{Zero-Shot Generation.} This baseline employs a direct synthesis approach where the LLM generates complete model cards from source papers and web repository pages in a single inference pass without retrieval augmentation. This method represents an end-to-end generation paradigm that processes entire documents holistically to produce comprehensive model documentation in one step.

\textbf{Template-Guided RAG (CardGen).} Following the established CardGen methodology \cite{liu2024automatic}, this baseline implements a structured query extraction approach using predefined question templates to systematically retrieve relevant information before generation. This method provides consistent structured extraction by employing a fixed set of queries designed to capture essential model card components across different paper types. Detailed hyperparameter configurations and implementation details for all models are provided in Appendix \ref{modeldetails}.

\textbf{Human-Annotated.} A detailed description of the human annotation process is provided in Section \ref{sec:benchmark}.

\subsubsection{Evaluation Metrics}

To comprehensively assess the quality of generated cards, we employ a five-dimensional evaluation framework. We adopt three established metrics from Liu et al. \cite{liu2024automatic}, \textit{Faithfulness}, \textit{Relevance}, and \textit{Accuracy}, which evaluate factual correctness and source alignment. However, these metrics cannot capture the structural coherence and practical utility requirements of multi-field documentation. Therefore, we introduce \textbf{two complementary metrics:} \textit{Consistency} measures internal coherence across independently generated fields to detect contradictions, while \textit{Usefulness} evaluates whether content provides actionable guidance for deployment decisions, addressing a critical gap in existing frameworks that assess correctness without considering operational value. Each metric is scored from 1 (poor) to 5 (excellent), and detailed definitions for all five metrics are provided in Table~\ref{tab:eval_metrics}.

\begin{table}[t]
\small
\centering
\caption{Evaluation metrics for assessing the generated model and data cards. Each metric is scored on a 1-5 scale by domain experts.}
\label{tab:eval_metrics}
\begin{tabular}{p{1.3cm}p{6.7cm}}
\toprule
\textbf{Metric} & \textbf{Definition} \\
\midrule
Faithfulness (F) & Accurately reflects information from source materials without introducing unsupported claims or omitting key points \\
\addlinespace
Relevance (R) & Content focused on the specific category being evaluated, avoiding unrelated or off-topic information \\
\addlinespace
Accuracy (A) & Statements are factually correct based on available references and can be directly verified \\
\addlinespace
Consistency (C) & Information is internally consistent within the card, with no contradictions or logical gaps \\
\addlinespace
Usefulness (U) & Provides clear, practical, and helpful information for users or researchers who want to understand or use the model or dataset \\
\bottomrule
\end{tabular}
\end{table}

\subsubsection{Evaluation Method}
We employ an LLM-as-a-Judge evaluation framework to assess generated card quality. This approach provides scalable, cost-effective assessment while maintaining high agreement with human judgments \cite{gu2024survey,li2024llms}, and is particularly well-suited for evaluating open-ended generation tasks where traditional metrics fail to capture nuanced qualities \cite{chiang2023can}. To mitigate potential biases inherent in any single large language model \cite{gu2024survey}, we utilize two distinct models as judges: GPT-5-nano and Gemini-2.5 Flash-Lite, and detailed evaluation procedures are provided in Appendix~\ref{appendix:eval_details}.

\subsection{Experimental Results}

Table~\ref{tab:llm_judge_performance} presents a comprehensive evaluation of \AQE \space relative to two baseline methods and the human-annotated MetaGAI-Bench across five quality dimensions, evaluated independently by two LLM judges.

\begin{table*}[!h]

\centering
\caption{Performance evaluation of LLM-as-a-Judge for model card and data card generation across quality dimensions. GPT = GPT-5-nano; Gemini = Gemini-2.5 Flash-Lite. Values are shown as Rank (Score). Correlation coefficients between GPT and Gemini: $\rho$ = Spearman correlation, $r$ = Pearson correlation. Significance levels: * $p<0.001$. Within each cell, \textbf{bold} indicates better performance (lower rank or higher score), \underline{underline} indicates the second-best performance.}
\label{tab:llm_judge_performance}
\resizebox{\textwidth}{!}{%
\begin{tabular}{ll|cc|cc|cc|cc|cc|cc}
\toprule
\multirow{2}{*}{\textbf{Card Type}} & \multirow{2}{*}{\textbf{Method}} & 
\multicolumn{2}{c|}{\textbf{Faithfulness}} & 
\multicolumn{2}{c|}{\textbf{Relevance}} & 
\multicolumn{2}{c|}{\textbf{Accuracy}} & 
\multicolumn{2}{c|}{\textbf{Consistency}} & 
\multicolumn{2}{c|}{\textbf{Usefulness}} & 
\multicolumn{2}{c}{\textbf{Average}} \\
& & 
\multicolumn{2}{c|}{$\rho$=0.215*, $r$=0.243*} & 
\multicolumn{2}{c|}{$\rho$=0.228*, $r$=0.375*} & 
\multicolumn{2}{c|}{$\rho$=0.204*, $r$=0.218*} & 
\multicolumn{2}{c|}{$\rho$=0.118*, $r$=0.141*} & 
\multicolumn{2}{c|}{$\rho$=0.331*, $r$=0.386*} & 
\multicolumn{2}{c}{} \\
\cmidrule(lr){3-4} \cmidrule(lr){5-6} \cmidrule(lr){7-8} \cmidrule(lr){9-10} \cmidrule(lr){11-12} \cmidrule(lr){13-14}
& & \textbf{GPT} & \textbf{Gemini} & \textbf{GPT} & \textbf{Gemini} & \textbf{GPT} & \textbf{Gemini} & \textbf{GPT} & \textbf{Gemini} & \textbf{GPT} & \textbf{Gemini} & \textbf{GPT} & \textbf{Gemini} \\
\midrule
Model & Zero-shot & 4.0 (4.02) & 3.0 (4.33) & 4.0 (4.64) & 3.0 (4.50) & 3.0 (4.07) & 3.0 (4.33) & \textbf{1.0 (4.75)} & \textbf{1.0 (4.66)} & 4.0 (3.72) & 3.0 (4.05) & 3.2 (4.24) & 2.6 (4.37) \\
Model & CardGen \cite{liu2024automatic} & 3.0 (4.07) & \underline{2.0 (4.36)} & \underline{2.0 (4.74)} & \underline{2.0 (4.62)} & \underline{2.0 (4.08)} & \underline{2.0 (4.36)} & 3.0 (4.73) & 3.0 (4.65) & \underline{2.0 (3.97)} & \underline{2.0 (4.19)} & \underline{2.4 (4.32)} & \underline{2.2 (4.43)} \\
Model & Human-Annotated & \textbf{1.0 (4.09)} & 4.0 (4.26) & 3.0 (4.66) & 4.0 (4.49) & \textbf{1.0 (4.11)} & 4.0 (4.26) & \underline{2.0 (4.75)} & 4.0 (4.62) & 3.0 (3.83) & 4.0 (4.04) & \textbf{2.0 (4.29)} & 4.0 (4.33) \\
Model & \AQE & \underline{2.0 (4.08)} & \textbf{1.0 (4.37)} & \textbf{1.0 (4.79)} & \textbf{1.0 (4.65)} & 4.0 (4.05) & \textbf{1.0 (4.37)} & 4.0 (4.73) & \underline{2.0 (4.65)} & \textbf{1.0 (4.02)} & \textbf{1.0 (4.24)} & \underline{2.4 (4.33)} & \textbf{1.2 (4.46)} \\
\midrule
Data & Zero-shot & 3.0 (4.10) & \underline{2.0 (4.27)} & 3.0 (4.62) & \underline{2.0 (4.57)} & \underline{2.0 (4.19)} & \underline{2.0 (4.27)} & \underline{2.0 (4.75)} & \underline{2.0 (4.67)} & 3.0 (3.83) & \underline{2.0 (4.08)} & 2.6 (4.30) & \underline{2.0 (4.37)} \\
Data & CardGen \cite{liu2024automatic} & 4.0 (4.01) & 4.0 (4.07) & 4.0 (4.57) & 4.0 (4.39) & 4.0 (4.07) & 4.0 (4.07) & 4.0 (4.67) & 4.0 (4.47) & 4.0 (3.82) & 4.0 (3.93) & 4.0 (4.23) & 4.0 (4.18) \\
Data & Human-Annotated & \underline{2.0 (4.12)} & 3.0 (4.19) & \underline{2.0 (4.64)} & 3.0 (4.51) & 3.0 (4.18) & 3.0 (4.19) & 3.0 (4.73) & 3.0 (4.60) & \underline{2.0 (3.86)} & 3.0 (4.05) & \underline{2.4 (4.31)} & 3.0 (4.31) \\
Data & \AQE & \textbf{1.0 (4.26)} & \textbf{1.0 (4.37)} & \textbf{1.0 (4.85)} & \textbf{1.0 (4.71)} & \textbf{1.0 (4.24)} & \textbf{1.0 (4.36)} & \textbf{1.0 (4.81)} & \textbf{1.0 (4.69)} & \textbf{1.0 (4.18)} & \textbf{1.0 (4.28)} & \textbf{1.0 (4.47)} & \textbf{1.0 (4.48)} \\
\bottomrule
\end{tabular}%
}
\end{table*}

\subsubsection{\textbf{Effective Integration of Web-Scale Information (RQ1)}}

\AQE \space effectively integrates heterogeneous web-scale information sources—scientific papers, repository metadata, and the MetaGAI Pool—to generate comprehensive documentation. For data cards, \AQE \space achieves first-place rankings across all dimensions for both judges (average scores: 4.47 GPT, 4.48 Gemini), substantially outperforming all baselines. For model cards, \AQE \space excels in Relevance (rank 1, scores: 4.79 GPT, 4.65 Gemini) and Usefulness (rank 1, scores: 4.02 GPT, 4.24 Gemini), with competitive performance on other metrics.

The differential performance across card types reveals the information integration challenges. Data card information follows more standardized documentation patterns, enabling systematic extraction, while model cards require nuanced interpretation of experimental details not always explicitly documented. Nevertheless, \AQE's automated approach achieves comparable or superior scores compared with human-annotated on most dimensions, validating multi-source web-scale integration.

\subsubsection{\textbf{Overcoming Static Template Limitations (RQ2)}}

Comparing \AQE \space and CardGen directly addresses RQ2. CardGen performs well for model cards (average rank 2.2-2.4) but poorly on data cards (rank 4), revealing a fundamental limitation of static templates: predefined questions capture standardized model properties but fail to accommodate diverse dataset characteristics.

\AQE's dynamic query expansion achieves dominant data card performance (rank 1 across all dimensions) while maintaining competitive model card results. Strong Relevance scores (4.79-4.85 for GPT) demonstrate effective information identification while filtering off-topic content. The performance gap on data cards (0.24-0.30 points) quantifies adaptive extraction's value over static templates for heterogeneous web documentation.

\subsubsection{\textbf{Multi-Dimensional Evaluation Framework (RQ3)}}

Our five-dimensional framework effectively assesses card quality, with each dimension capturing distinct aspects, as evidenced by varying inter-judge correlations ($\rho$=0.118-0.331, $r$=0.141-0.386).

\textbf{Faithfulness and Accuracy} evaluate content grounding. \AQE \space achieves first-rank performance on both metrics for data cards. For model cards, results show judge-dependent variation, with strong performance under Gemini evaluation (rank 1) but mixed results under GPT (Faithfulness rank 2, Accuracy rank 4). \textbf{Relevance} demonstrates strong discriminative power. \AQE's top rankings validate that this metric effectively captures whether iterative query expansion identifies pertinent information while filtering irrelevant content. \textbf{Consistency} reveals methodological tradeoffs. Zero-shot generation occasionally achieves top rankings, suggesting single-pass generation produces more uniform content, though potentially sacrificing completeness. Lower inter-judge correlation ($\rho$=0.118) indicates greater subjectivity. \textbf{Usefulness} assesses practical value beyond correctness. \AQE's top rankings demonstrate that dynamically retrieved content provides more actionable guidance than template-based or direct generation approaches.

\subsection{Human Evaluation}

\begin{table*}[!h]

\centering
\caption{Human evaluation of model card and data card generation across quality dimensions. Group 1 and Group 2 represent two independent sets of human evaluators assessing the same algorithms and card types. Values shown as Rank (Score). Correlation coefficients between Group 1 and Group 2: $\rho$ = Spearman correlation, $r$ = Pearson correlation. Significance levels: * $p<0.001$. Within each metric, \textbf{bold} indicates best performance (lowest rank or highest score), \underline{underline} indicates the second-best performance.}
\label{tab:human_evaluation}
\resizebox{\textwidth}{!}{%
\begin{tabular}{ll|cc|cc|cc|cc|cc|cc}
\toprule
\multirow{2}{*}{\textbf{Card Type}} & \multirow{2}{*}{\textbf{Method}} & 
\multicolumn{2}{c|}{\textbf{Faithfulness}} & 
\multicolumn{2}{c|}{\textbf{Relevance}} & 
\multicolumn{2}{c|}{\textbf{Accuracy}} & 
\multicolumn{2}{c|}{\textbf{Consistency}} & 
\multicolumn{2}{c|}{\textbf{Usefulness}} & 
\multicolumn{2}{c}{\textbf{Average}} \\
& & 
\multicolumn{2}{c|}{$\rho$=0.509*, $r$=0.467*} & 
\multicolumn{2}{c|}{$\rho$=0.204*, $r$=0.231*} & 
\multicolumn{2}{c|}{$\rho$=0.358*, $r$=0.366*} & 
\multicolumn{2}{c|}{$\rho$=0.310*, $r$=0.319*} & 
\multicolumn{2}{c|}{$\rho$=0.446*, $r$=0.453*} & 
\multicolumn{2}{c}{} \\
\cmidrule(lr){3-4} \cmidrule(lr){5-6} \cmidrule(lr){7-8} \cmidrule(lr){9-10} \cmidrule(lr){11-12} \cmidrule(lr){13-14}
& & \textbf{Group 1} & \textbf{Group 2} & \textbf{Group 1} & \textbf{Group 2} & \textbf{Group 1} & \textbf{Group 2} & \textbf{Group 1} & \textbf{Group 2} & \textbf{Group 1} & \textbf{Group 2} & \textbf{Group 1} & \textbf{Group 2} \\
\midrule
Model & Zero-shot & 4.0 (2.73) & 4.0 (2.94) & 4.0 (4.12) & 4.0 (3.01) & 4.0 (2.69) & 4.0 (2.85) & 4.0 (3.70) & 4.0 (3.17) & 4.0 (2.68) & 4.0 (2.79) & 4.0 (3.18) & 4.0 (2.95) \\
Model & CardGen \cite{liu2024automatic} & 3.0 (3.29) & 3.0 (3.21) & \underline{2.0 (4.53)} & 3.0 (3.33) & 3.0 (3.28) & 3.0 (3.17) & \underline{2.0 (4.19)} & 3.0 (3.39) & 3.0 (3.26) & 3.0 (3.09) & 2.6 (3.71) & 3.0 (3.24) \\
Model & Human-Annotated & \textbf{1.0 (3.69)} & \textbf{1.0 (3.53)} & \textbf{1.0 (4.66)} & \textbf{1.0 (3.62)} & \textbf{1.0 (3.66)} & \textbf{1.0 (3.43)} & \textbf{1.0 (4.32)} & \textbf{1.0 (3.59)} & \textbf{1.0 (3.55)} & \textbf{1.0 (3.35)} & \textbf{1.0 (3.98)} & \textbf{1.0 (3.50)} \\
Model & \AQE & \underline{2.0 (3.47)} & \underline{2.0 (3.31)} & 3.0 (4.38) & \underline{2.0 (3.43)} & \underline{2.0 (3.44)} & \underline{2.0 (3.22)} & 3.0 (4.05) & \underline{2.0 (3.45)} & \underline{2.0 (3.35)} & \underline{2.0 (3.17)} & \underline{2.4 (3.74)} & \underline{2.0 (3.32)} \\
\midrule
Data & Zero-shot & 4.0 (2.94) & 4.0 (2.76) & 4.0 (3.44) & 4.0 (2.41) & 4.0 (3.01) & 4.0 (2.13) & 4.0 (2.95) & 4.0 (2.20) & 4.0 (2.50) & 3.0 (2.23) & 4.0 (2.97) & 3.8 (2.35) \\
Data & CardGen \cite{liu2024automatic} & 3.0 (3.08) & 3.0 (2.98) & 3.0 (3.50) & 3.0 (2.61) & 3.0 (3.08) & 3.0 (2.25) & 3.0 (3.13) & 3.0 (2.53) & 3.0 (2.69) & 4.0 (2.21) & 3.0 (3.10) & 3.2 (2.52) \\
Data & Human-Annotated & \textbf{1.0 (3.79)} & \textbf{1.0 (4.01)} & \textbf{1.0 (3.95)} & \textbf{1.0 (3.92)} & \textbf{1.0 (3.57)} & \textbf{1.0 (3.57)} & \textbf{1.0 (3.43)} & \textbf{1.0 (3.56)} & \textbf{1.0 (3.22)} & \textbf{1.0 (3.45)} & \textbf{1.0 (3.59)} & \textbf{1.0 (3.70)} \\
Data & \AQE & \underline{2.0 (3.39)} & \underline{2.0 (3.15)} & \underline{2.0 (3.71)} & \underline{2.0 (2.83)} & \underline{2.0 (3.31)} & \underline{2.0 (2.60)} & \underline{2.0 (3.32)} & \underline{2.0 (2.70)} & \underline{2.0 (2.95)} & \underline{2.0 (2.42)} & \underline{2.0 (3.34)} & \underline{2.0 (2.74)} \\
\bottomrule
\end{tabular}%
}

\end{table*}

To complement the LLM-as-a-Judge assessments, we conducted an independent human evaluation study. We sampled 50 model cards and 50 data cards from MetaGAI-Bench. Six Data Science master's students, divided into two independent groups, evaluated all 100 cards across five quality dimensions using the 1-5 scoring rubric in Table~\ref{tab:eval_metrics}. Evaluators were selected from a different pool than benchmark annotators to prevent recognition bias, and results are presented in Table~\ref{tab:human_evaluation}.

We computed inter-group agreement using Spearman and Pearson correlation coefficients to assess evaluation consistency. Correlation coefficients vary by dimension: Faithfulness shows the highest agreement (\(\rho\)=0.509, \(r\)=0.467), followed by Usefulness (\(\rho\)=0.446, \(r\)=0.453), while Relevance exhibits the lowest (\(\rho\)=0.204, \(r\)=0.231). These moderate but statistically significant correlations (all \(p<0.001\)) indicate reasonable consistency, though variation reveals some quality aspects are more subjectively interpreted. Notably, human correlation values (\(\rho\)=0.204-0.509) exceed LLM judge correlations (\(\rho\)=0.118-0.331), suggesting more aligned human evaluation criteria.

\subsubsection{\textbf{Human Validation of Web-Scale Information Integration (RQ1)}}

Human evaluation validates that \AQE \space produces high-quality documentation approaching expert standards. \AQE \space consistently ranks second across card types and evaluator groups (model cards: rank 2.4/2.0; data cards: rank 2.0/2.0), substantially outperforming automated baselines while remaining competitive with human-annotated cards. 

Score gaps reveal remaining quality differences. For model cards, human-annotated achieve 3.53-3.66 versus \AQE's 3.35-3.44 (a gap of 0.18-0.22); for data cards, gaps widen to 0.40-0.85 points (human-annotated: 3.22-3.95 vs. \AQE: 2.42-3.71). This suggests current methods are less effective at capturing the explanatory depth and contextual appropriateness that human evaluators prioritize.

\subsubsection{\textbf{Human Perception of Dynamic vs. Static Approaches (RQ2)}}
Human evaluation confirms that dynamic query expansion overcomes static template limitations. \AQE \space outperforms CardGen across most quality dimensions, with notable advantages in Accuracy and Usefulness.

Performance gaps are more pronounced for data cards than model cards, mirroring LLM evaluation patterns. This convergent evidence supports the claim that dynamic adaptation is essential for diverse documentation structures where standardized templates fail, producing more actionable, contextually appropriate documentation.

\subsubsection{\textbf{Cross-Validation of Multi-Dimensional Evaluation \newline Framework (RQ3)}}
Human and LLM-based evaluation reveal complementary strengths. Human evaluators maintain higher inter-rater consistency (\(\rho\)=0.204-0.509) than LLM judges (\(\rho\)=0.118-0.331), but score systematically differently from each other: humans rate human-annotated cards higher, while scoring automated methods conservatively compared to LLMs' compressed distributions.

LLM judges prioritize syntactic correctness and factual accuracy, occasionally ranking \AQE \space first. Human evaluators apply holistic criteria encompassing explanatory depth, audience appropriateness, and practical utility. Convergence in relative rankings validates the framework's ability to distinguish method quality, while divergence in absolute scores reveals complementary evaluation perspectives.

\section{Conclusion}
\label{conclusion}
This work addresses the critical challenge of automating semantic documentation for GAI systems. We introduce the MetaGAI-Dataset, comprising 6,481 data cards and 123,013 model cards, along with a high-quality human-annotated MetaGAI-Bench of 600 data cards and 600 model cards. Additionally, we propose the \AQE \space framework that combines dynamic query expansion with cross-card knowledge transfer. Our analysis reveals significant documentation quality disparities, with data cards achieving substantially higher completeness than model cards and critical deficiencies in safety and ethical considerations. The proposed \AQE \space demonstrates that integrating dynamic retrieval with cross-card knowledge transfer enables scalable, high-quality semantic card generation, achieving first-place rankings across all quality dimensions for data cards and competitive performance for model cards in both LLM-based and human evaluations. Limitations include degraded performance for novel GAI architectures lacking reference cards and the inability to extract visual information from figures and tables. Future work will address multimodal knowledge extraction from figures and tables, unified data-model card mapping, and targeted methods for enhancing ethical documentation.

\bibliographystyle{ACM-Reference-Format}
\bibliography{references}

@article{gilbert2025could,
  title={Could transparent model cards with layered accessible information drive trust and safety in health AI?},
  author={Gilbert, Stephen and Adler, Rasmus and Holoyad, Taras and Weicken, Eva},
  journal={npj Digital Medicine},
  volume={8},
  number={1},
  pages={124},
  year={2025},
  publisher={Nature Publishing Group UK London}
}

@article{raza2025responsible,
  title={Who is responsible? the data, models, users or regulations? a comprehensive survey on responsible generative ai for a sustainable future},
  author={Raza, Shaina and Qureshi, Rizwan and Zahid, Anam and Kamawal, Safiullah and Sadak, Ferhat and Fioresi, Joseph and Saeed, Muhammaed and Sapkota, Ranjan and Jain, Aditya and Zafar, Anas and others},
  journal={arXiv preprint arXiv:2502.08650},
  year={2025}
}

@inproceedings{pushkarna2022data,
  title={Data cards: Purposeful and transparent dataset documentation for responsible ai},
  author={Pushkarna, Mahima and Zaldivar, Andrew and Kjartansson, Oddur},
  booktitle={Proceedings of the 2022 ACM Conference on Fairness, Accountability, and Transparency},
  pages={1776--1826},
  year={2022}
}

@inproceedings{liu2024automatic,
  title={Automatic Generation of Model and Data Cards: A Step Towards Responsible AI},
  author={Liu, Jiarui and Li, Wenkai and Jin, Zhijing and Diab, Mona},
  booktitle={Proceedings of the 2024 Conference of the North American Chapter of the Association for Computational Linguistics: Human Language Technologies (Volume 1: Long Papers)},
  pages={1975--1997},
  year={2024}
}

@inproceedings{mitchell2019model,
  title={Model cards for model reporting},
  author={Mitchell, Margaret and Wu, Simone and Zaldivar, Andrew and Barnes, Parker and Vasserman, Lucy and Hutchinson, Ben and Spitzer, Elena and Raji, Inioluwa Deborah and Gebru, Timnit},
  booktitle={Proceedings of the conference on fairness, accountability, and transparency},
  pages={220--229},
  year={2019}
}

@article{laufer2025anatomy,
  title={Anatomy of a Machine Learning Ecosystem: 2 Million Models on Hugging Face},
  author={Laufer, Benjamin and Oderinwale, Hamidah and Kleinberg, Jon},
  journal={arXiv preprint arXiv:2508.06811},
  year={2025}
}

@article{liang2024systematic,
  title={Systematic analysis of 32,111 AI model cards characterizes documentation practice in AI},
  author={Liang, Weixin and Rajani, Nazneen and Yang, Xinyu and Ozoani, Ezinwanne and Wu, Eric and Chen, Yiqun and Smith, Daniel Scott and Zou, James},
  journal={Nature Machine Intelligence},
  volume={6},
  number={7},
  pages={744--753},
  year={2024},
  publisher={Nature Publishing Group UK London}
}

@article{guo2024bias,
  title={Bias in Large Language Models: Origin, Evaluation, and Mitigation},
  author={Guo, Yufei and Guo, Muzhe and Su, Juntao and Yang, Zhou and Zhu, Mengqiu and Li, Hongfei and Qiu, Mengyang and Liu, Shuo Shuo},
  journal={CoRR},
  year={2024}
}

@article{heming2023benchmarking,
  title={Benchmarking bias: Expanding clinical AI model card to incorporate bias reporting of social and non-social factors},
  author={Heming, Carolina AM and Abdalla, Mohamed and Mohanna, Shahram and Ahluwalia, Monish and Zhang, Linglin and Trivedi, Hari and Woo, MinJae and Fine, Benjamin and Gichoya, Judy Wawira and Celi, Leo Anthony and others},
  journal={arXiv preprint arXiv:2311.12560},
  year={2023}
}

@article{semmelrock2025reproducibility,
  title={Reproducibility in machine-learning-based research: Overview, barriers, and drivers},
  author={Semmelrock, Harald and Ross-Hellauer, Tony and Kopeinik, Simone and Theiler, Dieter and Haberl, Armin and Thalmann, Stefan and Kowald, Dominik},
  journal={AI Magazine},
  volume={46},
  number={2},
  pages={e70002},
  year={2025},
  publisher={Wiley Online Library}
}

@inproceedings{mihalcea2025ai,
  title={Why AI Is WEIRD and Shouldn't Be This Way: Towards AI for Everyone, with Everyone, by Everyone},
  author={Mihalcea, Rada and Ignat, Oana and Bai, Longju and Borah, Angana and Chiruzzo, Luis and Jin, Zhijing and Kwizera, Claude and Nwatu, Joan and Poria, Soujanya and Solorio, Thamar},
  booktitle={Proceedings of the AAAI Conference on Artificial Intelligence},
  volume={39},
  number={27},
  pages={28657--28670},
  year={2025}
}

@article{longpre2024large,
  title={A large-scale audit of dataset licensing and attribution in AI},
  author={Longpre, Shayne and Mahari, Robert and Chen, Anthony and Obeng-Marnu, Naana and Sileo, Damien and Brannon, William and Muennighoff, Niklas and Khazam, Nathan and Kabbara, Jad and Perisetla, Kartik and others},
  journal={Nature Machine Intelligence},
  volume={6},
  number={8},
  pages={975--987},
  year={2024},
  publisher={Nature Publishing Group UK London}
}

@article{longpre2024data,
  title={Data Authenticity, Consent, \& Provenance for AI are all broken: what will it take to fix them?},
  author={Longpre, Shayne and Mahari, Robert and Obeng-Marnu, Naana and Brannon, William and South, Tobin and Gero, Katy and Pentland, Sandy and Kabbara, Jad},
  journal={arXiv preprint arXiv:2404.12691},
  year={2024}
}

@article{yang2024navigating,
  title={Navigating dataset documentations in AI: A large-scale analysis of dataset cards on hugging face},
  author={Yang, Xinyu and Liang, Weixin and Zou, James},
  journal={arXiv preprint arXiv:2401.13822},
  year={2024}
}

@article{bhardwaj2024state,
  title={The State of Data Curation at NeurIPS: An Assessment of Dataset Development Practices in the Datasets and Benchmarks Track},
  author={Bhardwaj, Eshta and Gujral, Harshit and Wu, Siyi and Zogheib, Ciara and Maharaj, Tegan and Becker, Christoph},
  journal={Advances in Neural Information Processing Systems},
  volume={37},
  pages={53626--53648},
  year={2024}
}

@article{puhlfurss2025model,
  title={Model Cards Revisited: Bridging the Gap Between Theory and Practice for Ethical AI Requirements},
  author={Puhlf{\"u}r{\ss}, Tim and Butzke, Julia and Maalej, Walid},
  journal={arXiv preprint arXiv:2507.06014},
  year={2025}
}

@article{sokol2024benchmarkcards,
  title={BenchmarkCards: Standardized Documentation for Large Language Model Benchmarks},
  author={Sokol, Anna and Daly, Elizabeth and Hind, Michael and Piorkowski, David and Zhang, Xiangliang and Moniz, Nuno and Chawla, Nitesh},
  journal={arXiv preprint arXiv:2410.12974},
  year={2024}
}

@article{comanici2025gemini,
  title={Gemini 2.5: Pushing the frontier with advanced reasoning, multimodality, long context, and next generation agentic capabilities},
  author={Comanici, Gheorghe and Bieber, Eric and Schaekermann, Mike and Pasupat, Ice and Sachdeva, Noveen and Dhillon, Inderjit and Blistein, Marcel and Ram, Ori and Zhang, Dan and Rosen, Evan and others},
  journal={arXiv preprint arXiv:2507.06261},
  year={2025}
}

@inproceedings{zhangbertscore,
  title={BERTScore: Evaluating Text Generation with BERT},
  author={Zhang, Tianyi and Kishore, Varsha and Wu, Felix and Weinberger, Kilian Q and Artzi, Yoav},
  booktitle={International Conference on Learning Representations},
  year={2019}
}

@article{cohen1960coefficient,
  title={A coefficient of agreement for nominal scales},
  author={Cohen, Jacob},
  journal={Educational and psychological measurement},
  volume={20},
  number={1},
  pages={37--46},
  year={1960},
  publisher={Sage Publications Sage CA: Thousand Oaks, CA}
}

@article{qwen3embedding,
  title={Qwen3 Embedding: Advancing Text Embedding and Reranking Through Foundation Models},
  author={Zhang, Yanzhao and Li, Mingxin and Long, Dingkun and Zhang, Xin and Lin, Huan and Yang, Baosong and Xie, Pengjun and Yang, An and Liu, Dayiheng and Lin, Junyang and Huang, Fei and Zhou, Jingren},
  journal={arXiv preprint arXiv:2506.05176},
  year={2025}
}

@article{gebru2021datasheets,
  title={Datasheets for datasets},
  author={Gebru, Timnit and Morgenstern, Jamie and Vecchione, Briana and Vaughan, Jennifer Wortman and Wallach, Hanna and Iii, Hal Daum{\'e} and Crawford, Kate},
  journal={Communications of the ACM},
  volume={64},
  number={12},
  pages={86--92},
  year={2021},
  publisher={ACM New York, NY, USA}
}

@article{panch2019artificial,
  title={Artificial intelligence and algorithmic bias: implications for health systems},
  author={Panch, Trishan and Mattie, Heather and Atun, Rifat},
  journal={Journal of global health},
  volume={9},
  number={2},
  pages={020318},
  year={2019}
}

@article{daneshjou2021lack,
  title={Lack of transparency and potential bias in artificial intelligence data sets and algorithms: a scoping review},
  author={Daneshjou, Roxana and Smith, Mary P and Sun, Mary D and Rotemberg, Veronica and Zou, James},
  journal={JAMA dermatology},
  volume={157},
  number={11},
  pages={1362--1369},
  year={2021},
  publisher={American Medical Association}
}

@inproceedings{jacovi2021formalizing,
  title={Formalizing trust in artificial intelligence: Prerequisites, causes and goals of human trust in AI},
  author={Jacovi, Alon and Marasovi{\'c}, Ana and Miller, Tim and Goldberg, Yoav},
  booktitle={Proceedings of the 2021 ACM conference on fairness, accountability, and transparency},
  pages={624--635},
  year={2021}
}

@article{gu2024survey,
  title={A survey on llm-as-a-judge},
  author={Gu, Jiawei and Jiang, Xuhui and Shi, Zhichao and Tan, Hexiang and Zhai, Xuehao and Xu, Chengjin and Li, Wei and Shen, Yinghan and Ma, Shengjie and Liu, Honghao and others},
  journal={arXiv preprint arXiv:2411.15594},
  year={2024}
}

@article{li2024llms,
  title={Llms-as-judges: a comprehensive survey on llm-based evaluation methods},
  author={Li, Haitao and Dong, Qian and Chen, Junjie and Su, Huixue and Zhou, Yujia and Ai, Qingyao and Ye, Ziyi and Liu, Yiqun},
  journal={arXiv preprint arXiv:2412.05579},
  year={2024}
}

@inproceedings{chiang2023can,
  title={Can Large Language Models Be an Alternative to Human Evaluations?},
  author={Chiang, Cheng-Han and Lee, Hung-Yi},
  booktitle={Proceedings of the 61st Annual Meeting of the Association for Computational Linguistics (Volume 1: Long Papers)},
  pages={15607--15631},
  year={2023}
}

@article{blecher2023nougat,
  title={Nougat: Neural optical understanding for academic documents},
  author={Blecher, Lukas and Cucurull, Guillem and Scialom, Thomas and Stojnic, Robert},
  journal={arXiv preprint arXiv:2308.13418},
  year={2023}
}

@article{arya2019one,
  title={One explanation does not fit all: A toolkit and taxonomy of ai explainability techniques},
  author={Arya, Vijay and Bellamy, Rachel KE and Chen, Pin-Yu and Dhurandhar, Amit and Hind, Michael and Hoffman, Samuel C and Houde, Stephanie and Liao, Q Vera and Luss, Ronny and Mojsilovi{\'c}, Aleksandra and others},
  journal={arXiv preprint arXiv:1909.03012},
  year={2019}
}

@article{phillips2021four,
  title={Four principles of explainable artificial intelligence},
  author={Phillips, P Jonathon and Phillips, P Jonathon and Hahn, Carina A and Fontana, Peter C and Yates, Amy N and Greene, Kristen and Broniatowski, David A and Przybocki, Mark A},
  year={2021},
  publisher={US Department of Commerce, National Institute of Standards and Technology}
}

@inproceedings{adkins2022prescriptive,
  title={Prescriptive and descriptive approaches to machine-learning transparency},
  author={Adkins, David and Alsallakh, Bilal and Cheema, Adeel and Kokhlikyan, Narine and McReynolds, Emily and Mishra, Pushkar and Procope, Chavez and Sawruk, Jeremy and Wang, Erin and Zvyagina, Polina},
  booktitle={CHI conference on human factors in computing systems extended abstracts},
  pages={1--9},
  year={2022}
}

@inproceedings{shen2021value,
  title={Value cards: An educational toolkit for teaching social impacts of machine learning through deliberation},
  author={Shen, Hong and Deng, Wesley H and Chattopadhyay, Aditi and Wu, Zhiwei Steven and Wang, Xu and Zhu, Haiyi},
  booktitle={Proceedings of the 2021 ACM conference on fairness, accountability, and transparency},
  pages={850--861},
  year={2021}
}

@inproceedings{seifert2019towards,
  title={Towards generating consumer labels for machine learning models},
  author={Seifert, Christin and Scherzinger, Stefanie and Wiese, Lena},
  booktitle={2019 IEEE First International Conference on Cognitive Machine Intelligence (CogMI)},
  pages={173--179},
  year={2019},
  organization={IEEE}
}

@article{brajovic2023model,
  title={Model reporting for certifiable ai: A proposal from merging eu regulation into ai development},
  author={Brajovic, Danilo and Renner, Niclas and Goebels, Vincent Philipp and Wagner, Philipp and Fresz, Benjamin and Biller, Martin and Klaeb, Mara and Kutz, Janika and Neuhuettler, Jens and Huber, Marco F},
  journal={arXiv preprint arXiv:2307.11525},
  year={2023}
}

@article{sidhpurwala2025blueprints,
  title={Blueprints of Trust: AI System Cards for End to End Transparency and Governance},
  author={Sidhpurwala, Huzaifa and Fox, Emily and Mollett, Garth and Gabarda, Florencio Cano and Zhukov, Roman},
  journal={arXiv preprint arXiv:2509.20394},
  year={2025}
}

@article{bender2018data,
  title={Data statements for natural language processing: Toward mitigating system bias and enabling better science},
  author={Bender, Emily M and Friedman, Batya},
  journal={Transactions of the Association for Computational Linguistics},
  volume={6},
  pages={587--604},
  year={2018},
  publisher={MIT Press One Rogers Street, Cambridge, MA 02142-1209, USA journals-info~…}
}

@article{holland2020dataset,
  title={The dataset nutrition label},
  author={Holland, Sarah and Hosny, Ahmed and Newman, Sarah and Joseph, Joshua and Chmielinski, Kasia},
  journal={Data protection and privacy},
  volume={12},
  number={12},
  pages={1},
  year={2020},
  publisher={Hart Publishing}
}

@article{roman2023open,
  title={Open datasheets: Machine-readable documentation for open datasets and responsible ai assessments},
  author={Roman, Anthony Cintron and Vaughan, Jennifer Wortman and See, Valerie and Ballard, Steph and Torres, Jehu and Robinson, Caleb and Ferres, Juan M Lavista},
  journal={arXiv preprint arXiv:2312.06153},
  year={2023}
}

@inproceedings{rondina2023completeness,
  title={Completeness of datasets documentation on ML/AI repositories: An empirical investigation},
  author={Rondina, Marco and Vetr{\`o}, Antonio and De Martin, Juan Carlos},
  booktitle={EPIA Conference on Artificial Intelligence},
  pages={79--91},
  year={2023},
  organization={Springer}
}

@article{horwitz2025charting,
  title={Charting and Navigating Hugging Face's Model Atlas},
  author={Horwitz, Eliahu and Kurer, Nitzan and Kahana, Jonathan and Amar, Liel and Hoshen, Yedid},
  journal={arXiv e-prints},
  pages={arXiv--2503},
  year={2025}
}

@article{bai2025autoschemakg,
  title={AutoSchemaKG: Autonomous Knowledge Graph Construction through Dynamic Schema Induction from Web-Scale Corpora},
  author={Bai, Jiaxin and Fan, Wei and Hu, Qi and Zong, Qing and Li, Chunyang and Tsang, Hong Ting and Luo, Hongyu and Yim, Yauwai and Huang, Haoyu and Zhou, Xiao and others},
  journal={arXiv preprint arXiv:2505.23628},
  year={2025}
}

@article{dagdelen2024structured,
  title={Structured information extraction from scientific text with large language models},
  author={Dagdelen, John and Dunn, Alexander and Lee, Sanghoon and Walker, Nicholas and Rosen, Andrew S and Ceder, Gerbrand and Persson, Kristin A and Jain, Anubhav},
  journal={Nature communications},
  volume={15},
  number={1},
  pages={1418},
  year={2024},
  publisher={Nature Publishing Group UK London}
}

@inproceedings{jiao2024text2db,
  title={TEXT2DB: Integration-Aware Information Extraction with Large Language Model Agents},
  author={Jiao, Yizhu and Li, Sha and Zhou, Sizhe and Ji, Heng and Han, Jiawei},
  booktitle={Findings of the Association for Computational Linguistics ACL 2024},
  pages={185--205},
  year={2024}
}

@inproceedings{suri2025visdom,
  title={VisDoM: Multi-Document QA with Visually Rich Elements Using Multimodal Retrieval-Augmented Generation},
  author={Suri, Manan and Mathur, Puneet and Dernoncourt, Franck and Goswami, Kanika and Rossi, Ryan A and Manocha, Dinesh},
  booktitle={Proceedings of the 2025 Conference of the Nations of the Americas Chapter of the Association for Computational Linguistics: Human Language Technologies (Volume 1: Long Papers)},
  pages={6088--6109},
  year={2025}
}

@article{papagiannidis2025responsible,
  title={Responsible artificial intelligence governance: A review and research framework},
  author={Papagiannidis, Emmanouil and Mikalef, Patrick and Conboy, Kieran},
  journal={The Journal of Strategic Information Systems},
  volume={34},
  number={2},
  pages={101885},
  year={2025},
  publisher={Elsevier}
}

@article{novelli2024accountability,
  title={Accountability in artificial intelligence: What it is and how it works},
  author={Novelli, Claudio and Taddeo, Mariarosaria and Floridi, Luciano},
  journal={Ai \& Society},
  volume={39},
  number={4},
  pages={1871--1882},
  year={2024},
  publisher={Springer}
}

@inproceedings{kroll2021outlining,
  title={Outlining traceability: A principle for operationalizing accountability in computing systems},
  author={Kroll, Joshua A},
  booktitle={Proceedings of the 2021 ACM Conference on fairness, accountability, and transparency},
  pages={758--771},
  year={2021}
}

\appendix
\section{Appendix}
\subsection{Definition of Model and Data Card}
\label{sec:definition}

The definitions of model and data cards for GAI are shown in Table \ref{tab:definition}.

\begin{table*}[htbp]
\small
\centering
\caption{Definitions of Model and Data Card for Generative AI}
\label{tab:definition}
\begin{tabular}{cp{4.0cm}p{11.7cm}}
\toprule
\textbf{Card} & \textbf{Field} & \textbf{Description} \\
\midrule
\multirow{8}{*}[-25pt]{\rotatebox[origin=c]{90}{\textbf{Model Card}}} & Model Details & Information about the model developer, architecture, size, training methodology, modalities, version, license, and contact details \\

 & Intended Use & Primary applications, target users, supported languages/domains, out-of-scope uses, and age restrictions \\

 & Generative Capabilities & Generation quality, content types, length limitations, consistency, latency, and customization options \\

 & Safety Considerations & Content safety measures, bias analysis, fairness metrics, red team testing, jailbreaking resistance, and child safety \\

 & Training Data & Training corpus details, data filtering processes, demographic representation, language coverage, consent/privacy, and evaluation datasets \\

 & Performance Metrics & Generation quality metrics, safety metrics, factual accuracy, bias metrics, cultural sensitivity, and robustness measures \\

 & Ethical Considerations & Dual-use risks, misinformation potential, intellectual property concerns, economic/environmental impact, cultural appropriation, privacy, and consent issues \\

 & Caveats \& Recommendations & Known limitations, deployment recommendations, monitoring requirements, and user guidelines \\
\hline
\multirow{12}{*}[-40pt]{\rotatebox[origin=c]{90}{\textbf{Data Card}}} & Dataset Details & Dataset name, version, creators/curators, funding, type, text language, license, and related resources \\

 & Dataset Structure & Instances, fields, missing information, relationships, splits, and size statistics \\

 & Data Collection & Collection process, data sources, timeframe, ethical review, consent process, and data validation \\

 & Data Processing & Preprocessing steps, cleaning procedures, labeling process, quality control, filtering criteria, and deduplication \\

 & Intended Use & Primary tasks, suitable/unsuitable applications, research applications, commercial applications, and prohibited uses \\

 & Bias \& Fairness & Demographic representation, geographic/temporal coverage, known biases, bias mitigation, and fairness considerations \\

 & Privacy \& Security & Personally identifiable information, sensitive information, privacy protection measures, data security, anonymization/pseudonymization, and retention/deletion policies \\

 & Content Analysis & Content types, harmful content identification, content moderation, toxicity analysis, misinformation risks, and cultural sensitivity \\

 & Legal \& Ethical & Copyright considerations, terms of use, ethical guidelines, compliance requirements, subject rights, and institutional review \\

 & Maintenance \& Updates & Maintenance plan, update frequency, versioning, error reporting, community contribution, and deprecation plan \\

 & Distribution \& Access & Access mechanism, distribution format, download instructions, API access, access restrictions, and citation requirements \\

 & Limitations \& Recommendations & Known limitations, recommended uses, usage guidelines, performance considerations, environmental impact, and future work \\
\bottomrule
\end{tabular}
\end{table*}

\subsection{Weighted Card Completeness Index}
\label{wcci}

The WCCI quantifies documentation quality through content availability and confidence levels. The metric incorporates three design principles: \textit{interpretability} for cross-artifact comparison; \textit{confidence-weighted evaluation} reflecting information reliability; and \textit{uniform field weighting} preventing systematic bias. For each field $i$ with content $c_i$ and confidence level $\text{conf}_i$, the completeness score is: 0.0 for missing fields, 1.0 for non-applicable fields, and $w(\text{conf}_i) \in \{0.25, 0.5, 0.75, 1.0\}$ for confidence levels \{low, medium, high, and certain\}. The overall WCCI is:

\begin{equation}
\text{WCCI} = \frac{1}{|\mathcal{F}|} \sum_{i \in \mathcal{F}} \text{completeness\_score}(i)
\end{equation}

Table~\ref{tab:correlations} shows correlations between WCCI scores and metadata tags.

\begin{table}[t] \centering \small \caption{Spearman Correlations between WCCI Scores and Metadata Tags. * indicates $p < 0.001$.} \label{tab:correlations} \begin{tabular}{@{}p{0.5cm}lp{2.5cm}p{1.5cm}p{1.5cm}} \toprule \textbf{Card Type} & \textbf{Category} & \textbf{Metadata Tags} & \textbf{Correlation ($\rho$)} & \textbf{Frequency (n)} \\ \midrule \multirow{12}{*}{\rotatebox{90}{\textbf{Model Cards}}}  & Popularity & likes & 0.408* & 123,013 \\ & & downloads & 0.285* & 123,013 \\  & Task & text-to-speech & 0.211* & 3,145 \\ & & text-to-audio & 0.173* & 2,781 \\ & & image-text-to-text & 0.132* & 2,872 \\ & & text-to-image & 0.105* & 1,223 \\ & & text-generation & -0.153* & 97,689 \\  & License & apache-2.0 & 0.374* & 12,520 \\ & & cc-by-nc-4.0 & 0.222* & 3,445 \\  & Framework & pytorch & 0.386* & 16,750 \\ \midrule \multirow{9}{*}{\rotatebox{90}{\textbf{Data Cards}}}  & Popularity & downloads & 0.257* & 6,481 \\ & & likes & 0.182* & 6,481 \\  & Task & zero-shot classification & 0.203* & 680 \\ & & text-classification & 0.201* & 1,220 \\ & & table-QA & 0.195* & 717 \\ & & question-answer & 0.178* & 3,261 \\ & & multiple-choice & 0.164* & 665 \\  & Annotation & crowdsourced & -0.384* & 1,389 \\ \bottomrule  \end{tabular} \end{table}

\subsection{Field-Level WCCI Analysis by Task Category}
\label{appendix:wcci_analysis}

\begin{figure}[t]
\centering

\begin{subfigure}[b]{0.48\textwidth}
    \centering
    \includegraphics[width=\textwidth]{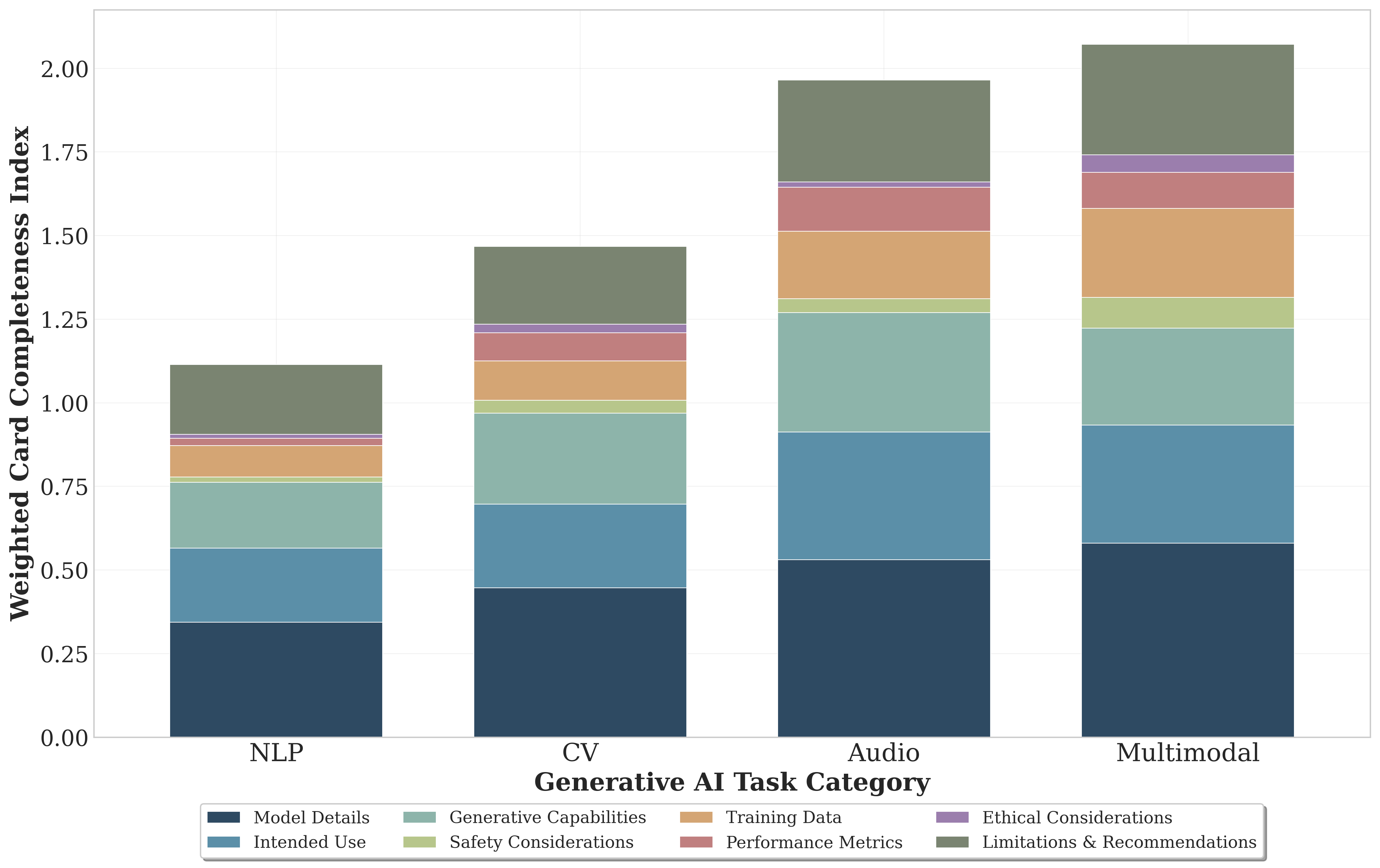}
    \caption{Model Cards}
    \label{fig:model_wcci}
\end{subfigure}
\hfill
\begin{subfigure}[b]{0.48\textwidth}
    \centering
    \includegraphics[width=\textwidth]{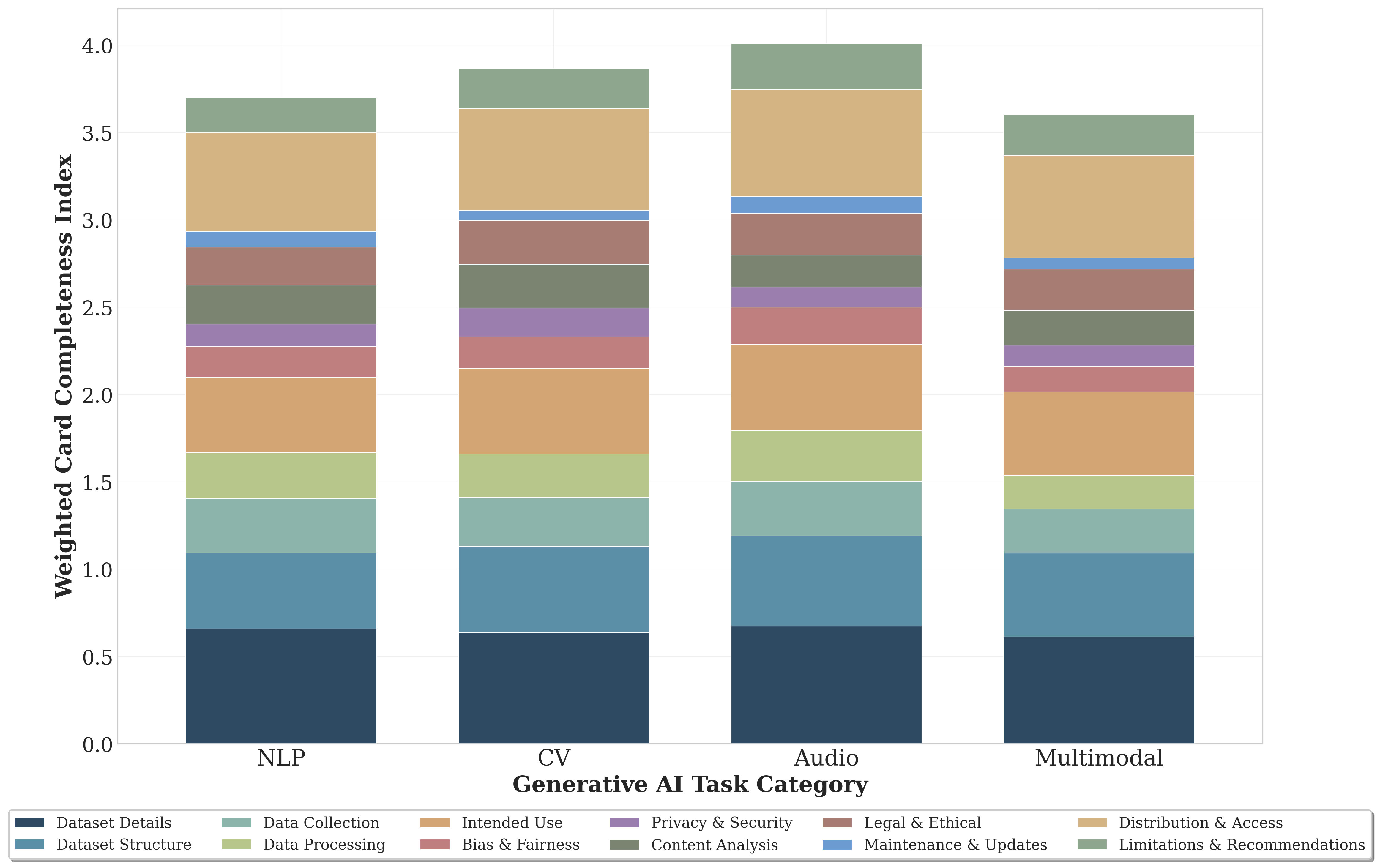}
    \caption{Data Cards}
    \label{fig:data_wcci}
\end{subfigure}
\caption{Field-level WCCI decomposition across generative AI task categories. (a) Model cards show substantially lower completeness with minimal Safety and Ethical Considerations. (b) Data cards achieve higher completeness dominated by technical specifications, with responsible AI dimensions contributing smaller segments.}
\label{fig:wcci_comparison}
\end{figure}

MetaGAI-Dataset exhibits pronounced documentation disparities across artifact types and modalities (Figure~\ref{fig:wcci_comparison}). Data cards substantially outperform model cards. Within model cards, performance varies by modality, with Multimodal and Audio exceeding CV and NLP. Critically, NLP demonstrates the lowest completeness despite representing 88\% of model cards. Field-level decomposition reveals that model cards contain negligible contributions from Safety Considerations, Ethical Considerations, and Performance Metrics. Data cards achieve higher completeness primarily through comprehensive technical documentation (Dataset Details, Dataset Structure, Data Collection), while responsible AI dimensions (Privacy \& Security, Legal \& Ethical, Limitations \& Recommendations) remain underrepresented across all modalities. This pattern indicates systematic prioritization of technical specifications over responsible AI documentation in both artifact types~\cite{yang2024navigating, liang2024systematic}.

\subsection{Model Details}
\label{modeldetails}
We employ Gemini-2.5 Flash-Lite \cite{comanici2025gemini} as the large language model for all card generation tasks. For fair comparison, both baseline methods (Zero-Shot and CardGen) also use Gemini-2.5 Flash-Lite as their underlying language model. We configure the model with temperature = 0.2, top-p sampling = 0.9, and maximum output tokens = 8,192. This configuration is consistently applied across all methods to ensure comparable evaluation conditions. For retrieval reranking, we employ the Qwen3-Reranker-4B model \cite{qwen3embedding}, configured with a maximum sequence length of 8,192 tokens, to process retrieved document chunks and score their relevance to queries.

\subsection{Evaluation Details}

\label{appendix:eval_details}

For LLM-as-a-Judge evaluation, we employ two distinct large language models: GPT-5-nano (version gpt-5-nano-2025-08-07 from OpenAI) and Gemini-2.5 Flash-Lite \cite{comanici2025gemini}. Each judge LLM independently assesses outputs from all algorithms on five metrics: Faithfulness, Relevance, Accuracy, Consistency, and Usefulness. To ensure robustness, we implement three bias mitigation strategies: (1) algorithm anonymization by removing all identifiers, (2) five independent evaluation rounds with randomized presentation orders, and (3) dual-judge assessment using two distinct LLMs. Final scores are computed as the average across both judges and all five rounds, yielding 10 independent assessments per evaluation instance.

\subsection{Ablation Study}

\subsubsection{MetaGAI Pool Contribution}
\begin{figure}[htbp]
    \centering
    \includegraphics[width=\columnwidth]{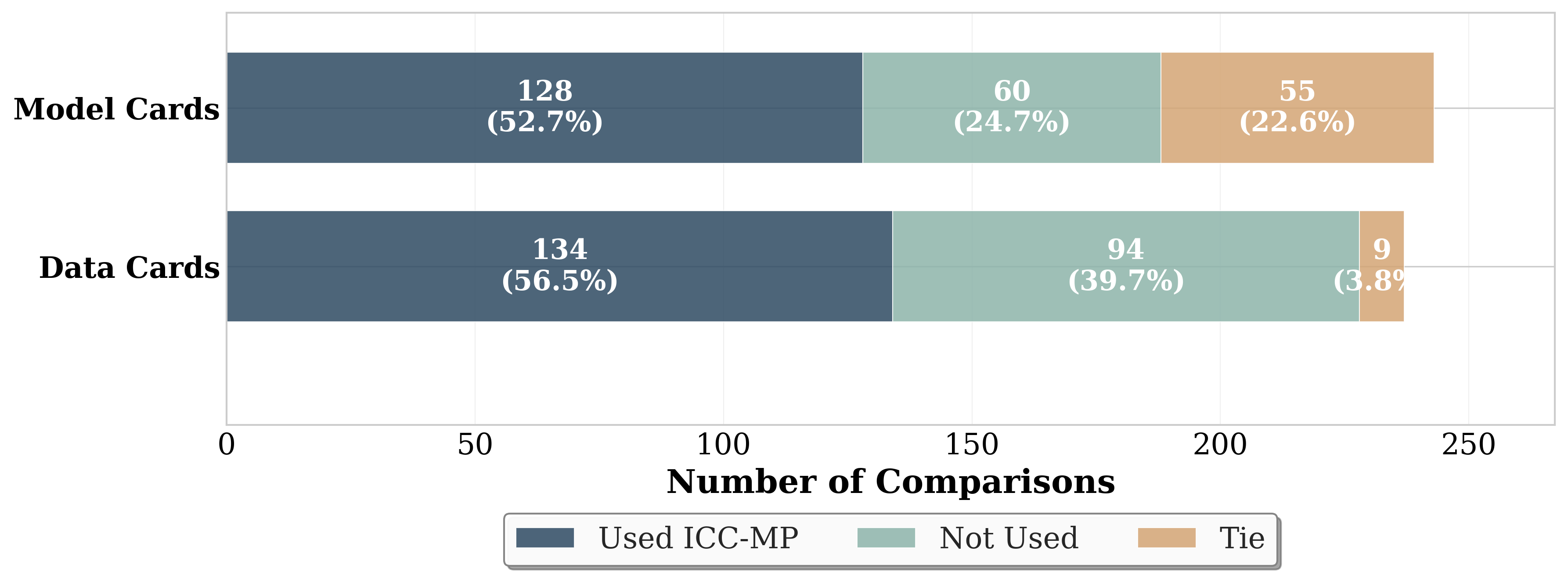}
    \caption{Quality improvement contribution of ICC-MP. Blue bars show cases where ICC-MP improved quality scores over IPE-QE-only cards, green bars show no improvement, and orange bars indicate tied scores.}
    \label{fig:metagai_contribution}
\end{figure}

ICC-MP successfully enhanced 52.7\% of model card fields and 56.5\% of data card fields (Figure~\ref{fig:metagai_contribution}). Of 243 model cards processed by ICC-MP, 128 (52.7\%) showed improvement, 60 (24.7\%) showed no change, and 55 (22.6\%) showed tied performance. Data cards demonstrated a stronger impact with 134 of 237 cards (56.5\%) improved, 94 (39.7\%) showing no change, and 9 (3.8\%) tied, validating cross-card knowledge transfer when source documents lack information.

\label{appendix:ablation}
\subsubsection{Effectiveness of Multi-Turn Query Expansion}
\begin{figure}[htbp]
    \centering
    \begin{subfigure}[b]{0.48\textwidth}
        \centering
        \includegraphics[width=\textwidth]{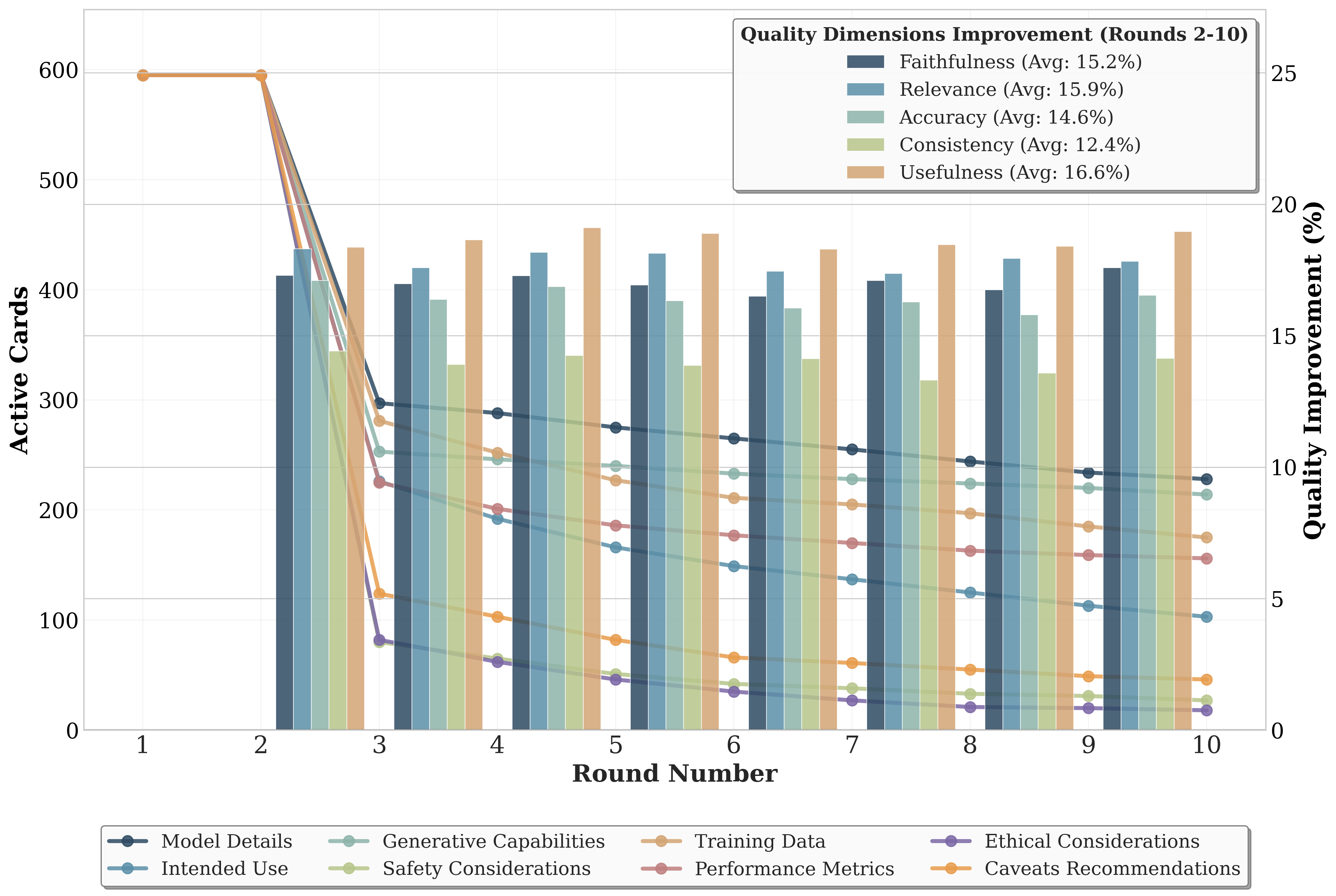}
        \caption{Model Cards}
        \label{fig:model-expansion}
    \end{subfigure}
    \hfill
    \begin{subfigure}[b]{0.48\textwidth}
        \centering
        \includegraphics[width=\textwidth]{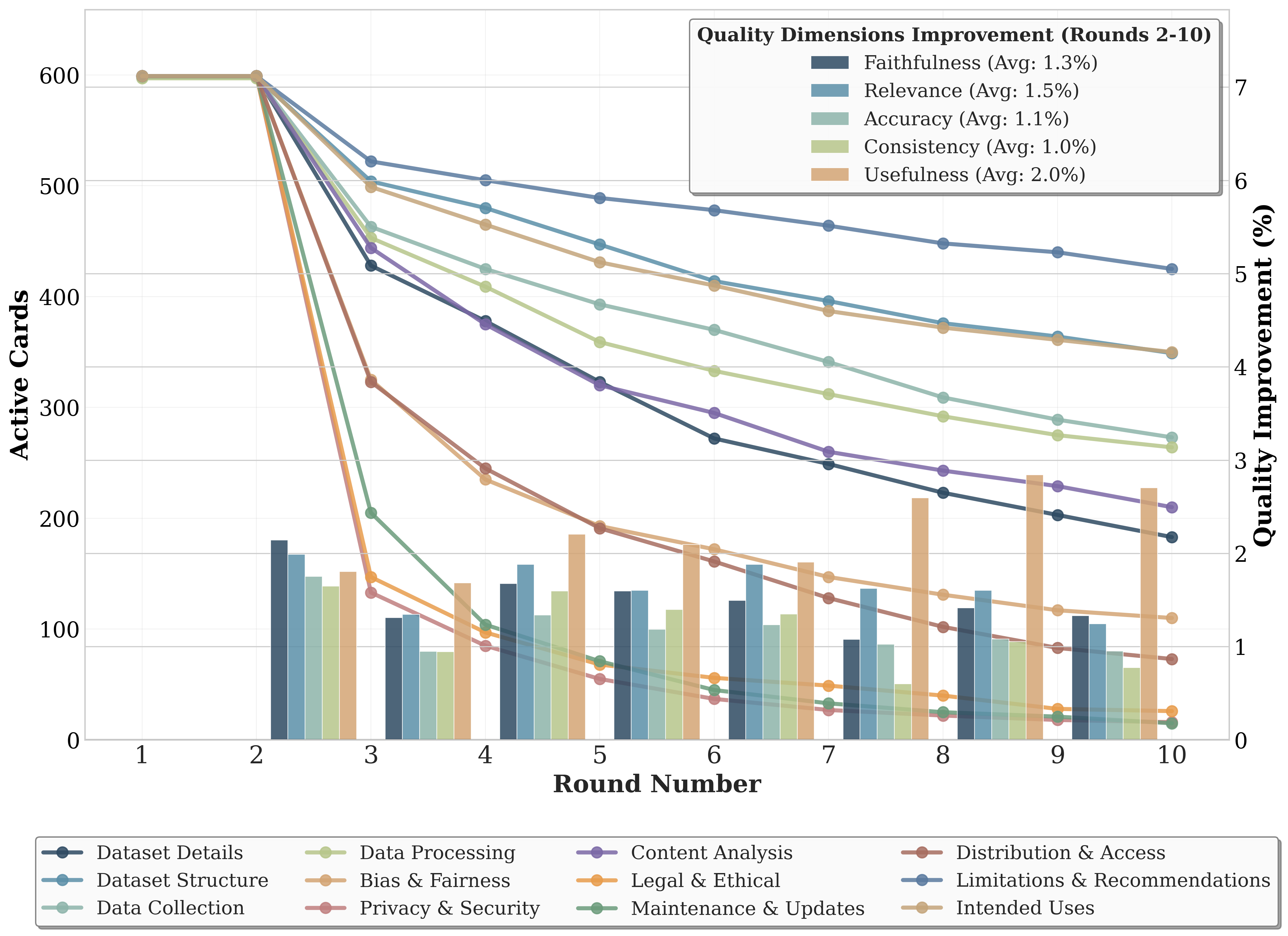}
        \caption{Data Cards}
        \label{fig:data-expansion}
    \end{subfigure}
    \caption{Query expansion effectiveness across rounds 2-10. Lines (left y-axis) show active cards undergoing query expansion per field; bars (right y-axis) track cumulative quality improvements across five dimensions.}
    \label{fig:query-expansion}
\end{figure}

Using LLM-as-a-Judge (Gemini-2.5 Flash-Lite), we compared consecutive rounds (\(R_{i-1}\) vs. \(R_i\)) across five quality dimensions. Figure~\ref{fig:query-expansion} reveals: (1) progressive convergence—active cards decline from 600 (Round 2) to 100-430 (Round 10); (2) sustained quality gains—model cards achieve 12-16\% average improvements per dimension while data cards show 1-2\% gains; (3) field-specific behavior—Model Details and Dataset Structure maintain higher activity (430-520 cards at Round 10), while responsible AI fields converge rapidly to under 200 cards.

\subsection{Case Study of Generated Model and Data Cards}
\label{casestudy}
To demonstrate \AQE's practical effectiveness, we present partial cards for \textit{all-hands/openhands-lm-32b-v0.1-ep3} (model) and \textit{MushanW/GLOBE\_V3} (data). IPE-QE extracted core information through iterative query refinement, while ICC-MP enriched incomplete fields using similar artifacts. Enrichments marked by \colorbox{yellow!30}{yellow background} and \textsuperscript{\textcolor{blue}{[enriched by ICC-MP]}} demonstrate successful cross-card knowledge transfer, validating \AQE's systematic extraction and conditional enrichment capabilities.

{\small
\begin{tcolorbox}[colback=gray!5, colframe=black, boxrule=1pt, rounded corners, title=\textbf{\href{https://huggingface.co/all-hands/openhands-lm-32b-v0.1-ep3}{Model Card: all-hands/openhands-lm-32b-v0.1-ep3}}, fonttitle=\bfseries]

\textbf{Model Details}\\
\vspace{2pt}
\begin{tabularx}{\linewidth}{@{}lX@{}}
\textit{Developer:} & OpenHands \\
\textit{Architecture:} & 32B Qwen2.5-Coder-Instruct. \\
\textit{License:} & \colorbox{yellow!30}{\textbf{MIT}} \textsuperscript{\textcolor{blue}{[enriched by ICC-MP]}} \\
... \\
\end{tabularx}

\vspace{4pt}
\textbf{Intended Use}\\
\vspace{2pt}
\begin{tabularx}{\linewidth}{@{}lX@{}}
\textit{Primary Applications:} & Agent scaffold for general-purpose prompting in software engineering tasks \\
\textit{Out of Scope:} & \colorbox{yellow!30}{\parbox{0.75\linewidth}{\textbf{Best suited for solving GitHub issues ...}}} \textsuperscript{\textcolor{blue}{[enriched by ICC-MP]}} \\
\end{tabularx}
...\\
\vspace{4pt}
\textbf{[Additional sections]}\\ 
\end{tcolorbox}
}

{\small
\begin{tcolorbox}[colback=gray!5, colframe=black, boxrule=1pt, rounded corners, title=\textbf{\href{https://huggingface.co/datasets/MushanW/GLOBE_V3}{Data Card: MushanW/GLOBE\_V3}}, fonttitle=\bfseries]

\textbf{Dataset Details}\\
\vspace{2pt}
\begin{tabularx}{\linewidth}{@{}lX@{}}
\textit{Dataset Name:} & GLOBE\_V3 \\
\textit{Version:} & \colorbox{yellow!30}{\parbox{0.73\linewidth}{\textbf{V3 version...}}} \textsuperscript{\textcolor{blue}{[enriched by ICC-MP]}} \\
...\\
\end{tabularx}

\textbf{Dataset Structure}\\
\vspace{2pt}
\begin{tabularx}{\linewidth}{@{}lX@{}}
\textit{Instances:} & The dataset includes utterances from 23,519 speakers. \\
\textit{Fields:} & Detailed metadata is available for all speakers ... \\
...
\end{tabularx}
\vspace{4pt}
\textbf{[Additional sections]}\\ 
\end{tcolorbox}
}

\end{document}